\documentclass[11pt, onecolumn, draft]{IEEEtran}
\usepackage{amsmath, amsfonts, amssymb, cite, epsfig, mathtools,adjustbox}
\usepackage{algorithm, algpseudocode, amsbsy,bm,bbm,amsthm, caption}
\usepackage{xcolor}
\usepackage{adjustbox,multirow, booktabs, hhline,tabularx, multirow}
\usepackage{subcaption}
\usepackage{dblfloatfix}

\makeatletter
\newcommand{\multiline}[1]{%
  \begin{tabularx}{\dimexpr\linewidth-\ALG@thistlm}[t]{@{}X@{}}
    #1
  \end{tabularx}
}

\newtheorem{theorem}{Theorem}
\newtheorem{lemma}{Lemma}
\newtheorem{definition}{Definition}

\newtheorem{assumption}{Assumption}

\DeclareMathOperator*{\argmin}{arg\,min}
\DeclareMathOperator*{\argmax}{arg\,max}

\usepackage{epstopdf}
\graphicspath{{figures/}}

\def \ExpFigWidth{0.7\textwidth}

\def\1{\mathbf 1}
\def\S{\mathcal S}

\begin{document}

\title{Context-Aware Online Client Selection for Hierarchical Federated Learning}
\author{Zhe~Qu,~Rui~Duan, Lixing~Chen, 
        Jie~Xu,~\IEEEmembership{Senior~Member,~IEEE,}
        Zhuo~Lu,~\IEEEmembership{Senior~Member,~IEEE}
        and~Yao~Liu,~\IEEEmembership{Senior~Member,~IEEE}
\thanks{Z. Qu, R. Duan and Z. Lu are with the Department
of Electrical Engineering, University of South Florida, Tampa,
FL, 33620 USA. E-mail: \{zhequ, ruiduan, zhuolu\}@usf.edu.}
\thanks{L. Chen is with the Institute of Cyber Science and Technology, Shanghai Jiao Tong University, Shanghai, Shanghai, 200240 China. E-mail:lxchen@sjtu.edu.cn.}
\thanks{J. Xu is with the Department of Electrical and Computer Engineering, University of Miami, 
Coral Gables, 33146 USA. E-mail: jiexu@miami.edu.}
\thanks{Y. Liu is with the Department of Computer Science and Engineering, University of South Florida, 
Tampa, 33620 USA. E-mail: yliu@cse.usf.edu.}}

\maketitle

\begin{abstract}
Federated Learning (FL) has been considered as an appealing framework to tackle data privacy issues of mobile devices compared to conventional Machine Learning (ML). Using Edge Servers (ESs) as intermediaries to perform model aggregation in proximity can reduce the transmission overhead, and it enables great potentials in low-latency FL, where the hierarchical architecture of FL (HFL) has been attracted more attention. Designing a proper client selection policy can significantly improve training performance, and it has been extensively used in FL studies. However, to the best of our knowledge, there are no studies focusing on HFL. In addition, client selection for HFL faces more challenges than conventional FL, e.g., the time-varying connection of client-ES pairs and the limited budget of the Network Operator (NO). In this paper, we investigate a client selection problem for HFL, where the NO learns the number of successful participating clients to improve the training performance (i.e., select as many clients in each round) as well as under the limited budget on each ES. An online policy, called Context-aware Online Client Selection (COCS), is developed based on Contextual Combinatorial Multi-Armed Bandit (CC-MAB). COCS observes the side-information (context) of local computing and transmission of client-ES pairs and makes client selection decisions to maximize NO's utility given a limited budget. Theoretically, COCS achieves a sublinear regret compared to an Oracle policy on both strongly convex and non-convex HFL. Simulation results also support the efficiency of the proposed COCS policy on real-world datasets.
\end{abstract}

\section{Introduction}\label{Sec:Introduction}
Federated Learning (FL) \cite{konevcny2016federated, mcmahan2017communication, li2019convergence, karimireddy2020scaffold} has become an attractive ML framework to address the growing concerns of transmitting private data from distributed clients (e.g., mobile devices) to a central cloud server by leveraging the ever-increasing storage and computing capabilities of the client devices. In each FL round, clients train local models using their local data and the cloud server aggregates local model updates to form a global model. Because only local model information is exchanged in FL rather than the local data, FL preserves the data privacy of the clients and hence has found applications in a wide range of problems, such as next-word prediction \cite{zhu2020empirical} and image classification \cite{guo2021multi}. 

A main bottleneck that limits the performance of FL is the delay variability among individual clients due to their local training and model data transfer via the wireless network. In standard FL, the cloud server has to wait until receiving the training updates from all the clients before processing any next step. Therefore, straggler clients who have unfavorable wireless links or low computation capabilities may dramatically slow down the whole FL process \cite{mao2017survey, yang2020federated}. This is the so-called ``straggler effect''. Various approaches have been proposed to mitigate the ``straggler effect''. For example, model quantization \cite{reisizadeh2020fedpaq, shlezinger2020uveqfed} and gradient sparsification \cite{sun2020adaptive} schemes aim to directly reduce the transferred data size and the model training complexity, thereby reducing all clients' training and transmission delay. Asynchronous FL \cite{wu2020safa, li2021stragglers} allows clients to train and upload training data in an asynchronous manner, and hence the cloud server does not have to wait for the slow clients to process the next step. Another mainstream and proven effective approach to address the straggler problem is client selection, which reduces the probability of straggler clients participating in FL by judiciously selecting clients in every FL round. In this paper, we aim to improve the FL performance along the lines of client selection in the context of hierarchical FL (HFL), which is a hierarchical architecture of FL that significantly reduces the communication overhead between the cloud server and the clients.

The key idea of HFL is the introduction of multiple edge servers which reside between the single cloud server and the large number of clients. Instead of communicating directly to the cloud server, the clients only send their local training updates to the nearby edge servers, each of which performs an intermediate model aggregation. These aggregated models are further sent to the cloud server for the global aggregation. HFL significantly reduces the communication burden on the network and has been shown to achieve a faster convergence speed than the conventional FL architecture both theoretically \cite{liu2020client, wang2020local} and empirically \cite{liu2021hierarchical}. Although several learning algorithms have been designed for HFL, simplifying assumptions have been made that all clients participate in each round of model parameter aggregation and hence, the ``straggler effect'' in HFL has not been specifically addressed. However, it is not straightforward to apply existing client selection solutions to HFL due to several unique challenges that HFL faces. Firstly, since the service area of an ES is much more restricted than Cloud Server (CS) and contains overlapping areas, the accessible clients of each ES are time-varying. This time-varying characteristic makes the client behavior of opportunistic communication more complicated, and Network Operator (NO) must carefully select the client to the corresponding ES in the overlapping area. Secondly, since the advantage of HFL is to deal with the straggler problem, how to design an efficient client selection policy is more important than traditional FL. Thirdly, the client selection decision needs to be determined based on many uncertainties in the HFL network conditions, e.g., the traffic pattern of client-ES pair and available computation resources of clients, which affect training performance in previously unknown ways. Therefore, a learning-based client selection policy is preferred to a solely optimization-based policy.

In this paper, we investigate the client selection problem for HFL and propose a new learning-based policy, called Context-aware Online Client Selection (COCS). COCS is developed based on a novel Multi-Armed Bandit (MAB) framework called Contextual Combinatorial MAB (CC-MAB) \cite{chen2018contextual, chen2019budget, chen2020learning}. COCS is contextual because it allows clients to use their computational information, e.g., available computation resources, and the client-ES pairs transmission information, e.g., bandwidth and distance. COCS is combinatorial because NO selects a subset of client-ES pairs and attempts to maximize the training utilities (i.e., select as many as clients in each round) by optimizing the client selection decision. To the best of our knowledge, COCS policy is the first client selection decision for HFL. In summary, we highlight the contributions of this paper as follows:

1) We formulate a client selection problem for HFL, where NO needs to select clients to ESs clients to process the local training in order to make more clients to be received by ESs before deadline under limited budget. To improve the convergence speed for HFL, client selection decision has a three-fold problem: (i) estimate the local model updates successfully received by ESs with cold-starts, (ii) decide whether a client should be selected to a certain ES due to time-varying connection conditions, and (iii) optimize how to pay computation resources on clients to maximize the utility under limited budgets.
    
2) Due to the a priori uncertain knowledge of participated clients, the client selection problem is formulated as a CC-MAB problem. An online learning policy called Context-aware Online Client Selection (COCS) is developed, which leverages the contextual information such as downloading channel state and local computing time over aggregation round for making a decision. For the strongly convex HFL, we analyze the utility loss of COCS, termed regret, compared to the Oracle solution that knows the exacted information of participated clients. A sublinear regret bound is derived for the proposed COCS policy, which implies that COCS can produce asymptotically optimal client selection decisions for HFL.
    
3) For non-convex HFL, the utility function of the convergence speed is quadratically related to the number of participated clients. By assuming that the information of each client-ES pair is perfectly known by NO, we show that the client selection problem is a submodular maximization problem with $M$ knapsack and one matroid constraints, where $M$ is the number of ESs. We use the Fast Lazy Greedy (FLGreedy) algorithm \cite{badanidiyuru2014fast} to approximate the optimal solution with a performance guarantee. To this end, the analysis shows that the COCS policy also achieves a sublinear regret.

The rest of this paper is organized as follows: Section~\ref{Sec:Related} overviews the related works. The system model and client selection problem of HFL are presented in Section~\ref{Sec:System}. We design the Context-aware Online Client Selection (COCS) policy for strongly convex HFL and provide an analytical performance guarantee in Section~\ref{Sec:COCS}. Section~\ref{Sec:Extension} presents the COCS policy for non-convex HFL, which is applied by the approximated oracle solutions. Simulation results are shown in Section~\ref{Sec:Experiments}, followed by the conclusion in Section~\ref{Sec:Conclusion}.

\section{Related Work}\label{Sec:Related}
We provide a brief literature review that includes three main lines of work: client selection for FL, HFL, and the MAB problems for FL.

Due to the heterogeneous clients and limited resources of wireless FL networks, client selection can significantly improve the performance of FL in terms of convergence speed and training latency. For example, \cite{wang2020optimizing} designs a deep reinforcement learning algorithm (the local model updates and the global model are considered as states) to select clients. \cite{nguyen2020fast} uses gradient information to select clients. If the inner product between the client's local and global gradient is negative, it will be excluded. In \cite{shi2020communication}, they develop a system model to estimate the total number of aggregation rounds and design a greedy algorithm to jointly optimize client selection and bandwidth allocation throughout the training process, thereby minimizing the training latency. \cite{xu2020client} designs a dynamic client sampling method. In the early aggregation rounds, fewer clients are selected, and more clients in later rounds. This method has been proven to improve the training loss and accuracy as well as decreasing the overall energy consumption. These client selection methods focus on the conventional FL, which is different from our consideration of HFL.

HFL has been considered to be a more practical FL framework for the current MEC system, since the hierarchical architecture makes FL communication more efficient and significantly reduces the latency \cite{liu2020client}. Later, some studies improve the performance of HFL from different perspectives or use it in some other applications. For example, \cite{liu2021hierarchical, wang2020local} propose a detailed convergence analysis of HFL, showing that the convergence speed of HFL achieves a linear speedup of conventional FL. Recently, FL has attracted the more interest, especially with the rapid development of ML applications on IoT devices. \cite{chai2020hierarchical} designs a hierarchical blockchain framework for knowledge sharing on smart vehicles, which learn the environmental data through ML methods and share the learning knowledge with others. \cite{wu2021hierarchical} uses HFL to better adapt to personalized modeling tasks and protect private information. 

The MAB problem has been extensively studied to address the key trade-off between exploration and exploitation making under uncertain environment \cite{auer2002finite}, and it has been used in FL for designing the client scheduling or selection \cite{xia2020multi, huang2020efficiency, wei2021low}. \cite{xia2020multi} designs a client scheduling problem and provides a MAB-based framework for FL training without knowing the wireless channel state information and the dynamic usage of local computing resources. In order to minimize the latency, \cite{huang2020efficiency} models fair-guaranteed client selection as a Lyapunov optimization problem and presents a policy based on CC-MAB to estimate the model transmission time. A multi-agent MAB algorithm is developed to minimize the FL training latency over wireless channels, constrained by training performance as well as each client's differential privacy requirement in \cite{wei2021low}. In this paper, the COCS policy is proposed to select clients for HFL. Due to the dynamic connection conditions of the client-ES pair and the limited available computing capacities of clients in each edge aggregation round, the COCS policy faces more challenges than these studies.

\section{System Model and Problem Formulation}\label{Sec:System}
\subsection{Preliminary of Hierarchical Federated Learning}
The Network Operator (NO) leverages a typical edge-cloud architecture to offer the Federated Learning (FL) service, known as Hierarchical FL (HFL) \cite{liu2020client, liu2021hierarchical, wang2020local} as in Fig.~\ref{Fig:HFL}. Unlike the conventional FL \cite{mcmahan2017communication,konevcny2016federated, li2019convergence, karimireddy2020scaffold} that involves only clients and a Cloud Server (CS), HFL consists of a set of mobile devices/clients, indexed by $\mathcal{N}=\{1,2,\dots,N\}$, a set of Edge Servers (ES), indexed by $\mathcal{M}=\{1,2,\dots,M\}$ and a Cloud Server (CS). Let $\mathcal{N}_m^t = \{1,2,\dots,N_m^t \}$ denote the set of clients that can communicate with ES $m$ in edge aggregation in round $t$. Note that the communication areas of different edge servers may be overlapped (i.e., $\sum_{m=1}^M N_m^t \geq N$). For each client $n \in \mathcal{N}$, it is able to communicate with a subset of ESs $\mathcal{C}^t_n \subseteq \mathcal{M}$ in round $t$. In particular, we assume that each client is equipped with a single antenna so that it can communicates with only one ES $m \in \mathcal{C}_n^t$ at a time even if it is located in the overlapped area of multiple ESs in each round. Let $\bm{w}$ denote the parameters of the global model. The goal of the FL service is to find the optimal parameters of the global model $\bm{w}^*$, which minimizes the average loss function $f(\bm{w})$ under the HFL network as follows:
\begin{equation}\label{Eq:FL}
    \min_{\bm{w}}f(\bm{w}) := \frac{1}{M}\sum\nolimits_{m\in \mathcal{M}}\frac{1}{S_m}\sum\nolimits_{n \in \bm{s}_m}F_n (\bm{w}),
\end{equation}
where $\bm{s}_m$ is the selected client set by the ES $m$ with the number $S_m$ in each round, $F_n (\bm{w}) \triangleq \sum_{\xi_n \sim \mathcal{D}_n} \ell (\bm{w} ; \xi_n )$ is the loss function associated with the local dataset $\mathcal{D}_n$ on client $n$, and $\ell(\bm{w};\xi_n )$ is the loss of data sample $\xi_n$. The objective of the loss function $F_n (\cdot)$ can be convex, e.g., logistic regression, or non-convex, e.g., neural network. The training steps of HFL can be summarized as follows:

\begin{figure}[t!]
    \centering
    \includegraphics[draft=false,width=\ExpFigWidth]{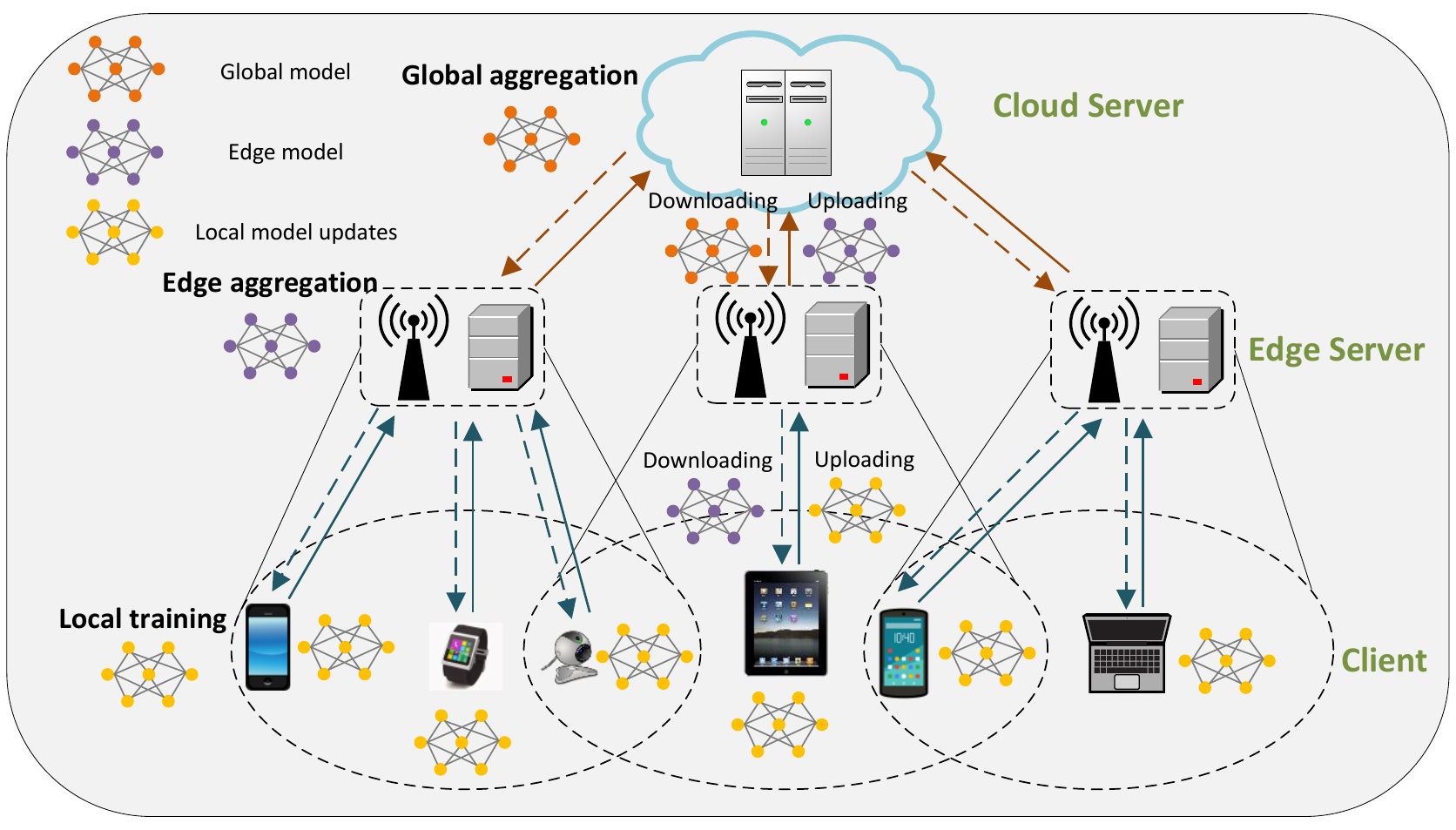}
    \caption{The architecture of HFL.}
    \label{Fig:HFL}
\end{figure}

(i) At the beginning of round $t$, each ES $m$ randomly selects a subset of clients $\bm{s}_m^t \subseteq \mathcal{N}_m^t$ in its coverage area. Even if a client is in the overlapping area, is is only allowed to communicate with one ES in one round. We assume that the HFL network contains a backhaul link to transmit the selected clients to avoid that some clients are selected on multiple ESs. Each client $n$ selected by ES $m$ downloads the edge model $\bm{w}_m^t$ and sets it to be the local model $\bm{w}_{n}^t = \bm{w}_m^t$, $\forall n \in \bm{s}_m^t$.

(ii) Then, each client $n$ takes $E$ epochs to update its own local model by Stochastic Gradient Descent (SGD) from its dataset $\mathcal{D}_n$ as follows:
\begin{equation}\label{Eq:SGD}
    \bm{w}_{n}^{t+e+1} = \bm{w}_{n}^{t+e} - \eta_t g (\bm{w}_{n}^{t+e} ; \xi_{n}^{t+e}),    
\end{equation}
where $e = 0,1,\dots, E-1$, $\eta_t$ is the learning rate, and $g(\bm{w}_n^{t+e};\xi_n^{t+e})$ is the stochastic gradient of $F_n (\bm{w})$ (i.e., $\mathbb{E}_{\xi_n^{t+e} \sim \mathcal{D}_n}[g(\bm{w}_n^{t+e};\xi_n^{t+e})] = \nabla F_n (\bm{w}_n^{t+e})$).

(iii) After $E$ local training epochs, client $n \in \bm{s}_m^t$ uploads the local model updates $\Delta_{n}^t \triangleq \bm{w}_{n}^{t+E-1} - \bm{w}_n^t$ to the ES $m$. Instead of aggregating all local models on CS at the end of round $t$ \cite{mcmahan2017communication, li2019convergence, karimireddy2020scaffold} of conventional FL, local model updates are averaged within ES $m$ to be edge model $\bm{w}_m^{t+1}$, called \textit{edge aggregation}, which is given as follows:
\begin{equation}\label{Eq:edgeaggregation}
    \bm{w}_m^{t+1} = \bm{w}_m^t + \frac{1}{S_m^t}\sum\nolimits_{n \in \bm{s}_m^t}\Delta_{n}^t,
\end{equation}

(iv) Every $T_{ES}$ rounds of edge aggregation, global model $\bm{w}^t$ is computed by $\bm{w}^t = \frac{1}{M}\sum\nolimits_{m=1}^M \bm{w}_m^t$, $\forall t = \{T_{ES}, 2T_{ES}, \dots\}$ from all $M$ ESs, called \textit{global aggregation}. Then, each ES $m$ downloads the global model to be its edge model $\bm{w}_m^{t} = \bm{w}^t$.

Repeating the above four steps with a sufficiently large number $T$ of rounds, NO will achieve the global model $\bm{w} = \bm{w}^T$ and stop the training process. HFL has been demonstrated to be able to achieve a linear speedup of convergence compared to conventional FL algorithms \cite{liu2021hierarchical, wang2020local}.

\subsection{Cost of Client Selection}
Since clients usually do not belong to NO, clients charge NO for the amount of requested computation resources for collecting the dataset and performing the local training to achieve the learning goal. At the beginning of each edge aggregation round, each client reveals its available computation resources $y_n^t$, including CPU frequency, RAM and storage for the current round $t$. NO pays $c_n(y^t_n)$ to the client depending on the available computation resources, where $c_n (\cdot)$ is a non-decreasing function. Due to the limited budget $\tilde{B}$ of NO, in any edge aggregation round $t$, the client selection decision of NO must satisfy the budget constraint $\sum\nolimits_{m\in\mathcal{M}}\sum\nolimits_{n\in\bm{s}_m^t} c_n (y_n^t )\leq \tilde{B}$.


\subsection{Deadline Based HFL}\label{SubSec:deadline}
In summary, an edge aggregation consists of four stages: \textit{Download Transmission} (DT), \textit{Local Computation} (LC), \textit{Upload Transmission} (UT) and \textit{Edge Computation} (EC). 

In DT stage, the selected client $n \in \bm{s}_m^t$ downloads the current edge model from the ES $m$. Followed by Shannon's equation, the channel state of DT $c^t_{\text{DT},n}$ is calculated by: 
\begin{equation}\label{Eq:DTrate}
c^t_{\text{DT}, n} = \log_2 (1+ P^t_n g^t_{\text{DT},n} / N_0),
\end{equation}
where $P^t_n$ is the transmission power, $g^t_{\text{DT},n}$ is the downlink wireless channel gain and $N_0$ is the noise power. Let $a_{\text{DT}}$ denote the downloading data size (i.e., size of edge model $\bm{w}_m^t$) and the allocated bandwidth is $b_n^t$ in the edge aggregation round $t$. Therefore, thus the DT time for client $n$ is $\tau^t_{\text{DT},n} = a_{\text{DT}}/(b_n^t c^t_{\text{DT},n})$. Once the client $n$ receives $\bm{w}^t_m$, training comes to the LC stage (i.e., it updates the local model using its own dataset $\mathcal{D}_n$ according to Eq.~\eqref{Eq:SGD}). The LC time of each client is determined by the local computation resources $y_n^t$ in the current round $t$. Given the computation resources $y_n^t > 0$, the LC time can be obtained as $\tau_{\text{LC},n}^t (y_n^t ) = q/ y_n^t$, where $q$ is the computation workload, which is based on the complexity of learning model and data. When the LC is finished, client $n$ uploads its local model updates $\Delta_n^{t+1}$ to the ES $m$. Similar to the channel state definition of DT in Eq.~\eqref{Eq:DTrate}, the channel state of UT is $c^t_{\text{UT},n} =  \log_2 (1+P^t_n g^t_{\text{UT},n} /N_0 )$ and UT time is $\tau^t_{\text{UT},n}= a_{\text{UT}}/(b_n^t r^t_{\text{UT},n})$, where $g^t_{\text{UT},n}$ is the uplink channel gain and $a_{\text{UT}}$ is the uploading data size (i.e., size of $\Delta_n^{t+1}$). Finally, if the local model updates of all selected clients are received by ESs, the edge models should be computed according to Eq.~\eqref{Eq:edgeaggregation}. The EC time is $\tau^t_{\text{EC},m} = q_m / y_m$, where $q_m$ is the edge model workload and $y^m$ is the process capacity of the ES $m$. Since $\tau_{\text{EC},m}$ is the same for all the selected clients at ES $m$, the training time of client $n$ is defined as follows:
\begin{equation}\label{Eq:time}
\begin{split}
    \tau^t_n (y_n^t ) & = \tau^t_{\text{DT}, n} + \tau^t_{\text{LC},n}(y_n^t ) + \tau^t_{\text{UT},n} \\
    &= \frac{a_{\text{DT},n}^t}{b_n^t c_{\text{DT},n}^t} + \frac{q}{y_n^t}+ \frac{a_{\text{UT},n}^t}{b_n^t c_{\text{UT},n}^t}, ~~~ \forall n,t.
\end{split}
\end{equation}

Due to some physical limitations, e.g., low computation capability and unstable communication, some clients may incur huge training latency in one edge aggregation round. Therefore, the deadline-based FL \cite{nishio2019client, tran2019federated, yang2021characterizing} is more realistic to deal with straggler clients. Specifically, ESs drop the clients whose the local model updates cannot be received before the deadline $\tau_{\text{dead},m}$ (i.e., client $n$ such that $\tau^t_n > \tau_{\text{dead},m}$). In this paper, we consider deadline-based HFL. Therefore, the edge aggregation can be reformulated:
\begin{equation}\label{Eq:deadlinecount}
\bm{w}^t_m = \left\{
\begin{array}{ll}
     \frac{1}{\sum\nolimits_{n\in\bm{s}^t_m}X^t_n}\sum\nolimits_{n \in \bm{s}^t_m}X^t_n \bm{w}_n^t, &\text{if}~\sum_{n \in \bm{s}^t_m} X^t_n \geq Z \\
     \frac{1}{Z}\sum\nolimits_{\tau^{t}_n \leq \tau^{Z}}\bm{w}_n^t, &\text{else}
\end{array}\right.
\end{equation}
where $X_n^t$ is a binary random variable representing whether client $n$'s model update can be received before the deadline (i.e., if $\tau^t_n \leq \tau_{\text{dead},m}$, $X_n^t = 1$; otherwise, $X_n^t = 0$), and $\tau^Z$ is the training time of the $Z$-th fastest client. In order to guarantee a minimum level of training performance, we require that at least $Z$ local model updates must be used for edge aggregation. Therefore, in case less than $Z$ clients' updates are received before the deadline, the system has to wait for some additional time $\tau^Z - \tau_{\text{dead}, m}$. For practical values of $\tau^Z - \tau_{\text{dead}, m}$, the probability of having less than $Z$ client updates received before the deadline is small. For analysis convenience, we assume that at least $Z$ client updates can be received before the deadline in every edge aggregation round. In addition, we assume that the deadline of all ESs are set the same, $\tau_{\text{dead},m} = \tau_{\text{dead}}, \forall m$. The extension to heterogeneous deadlines is straightforward.

\begin{figure}[t!]
  \centering
  \begin{minipage}{0.48\columnwidth}
      \includegraphics[draft=false,width=\linewidth, height=0.7\columnwidth]{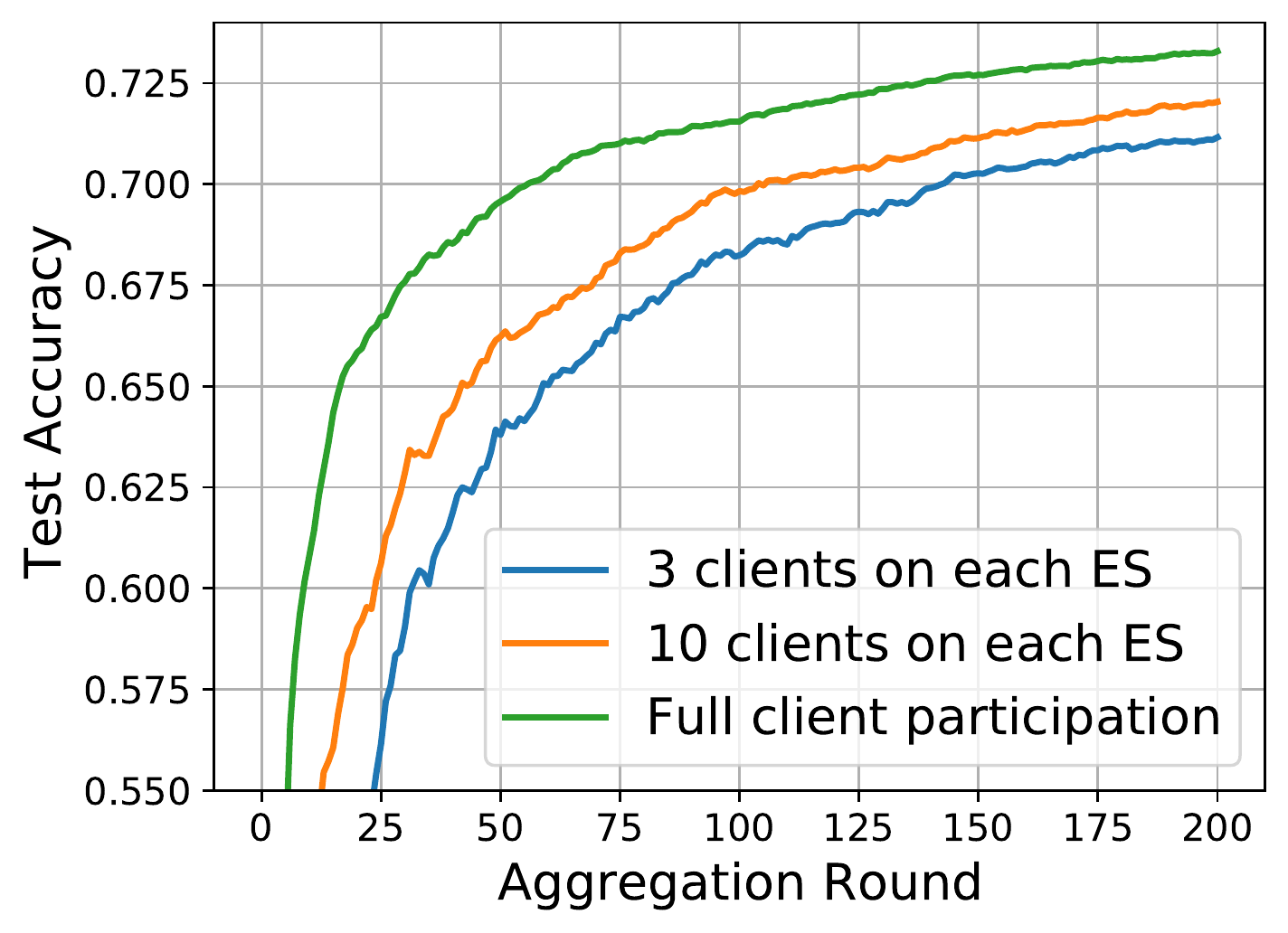}
      \subcaption{MNIST dataset under logistic regression.}
      \label{Fig:MNISTdiff}
    \end{minipage}
    \hfill
    \begin{minipage}{0.48\columnwidth}
      \includegraphics[draft=false,width=\linewidth, height=0.7\columnwidth]{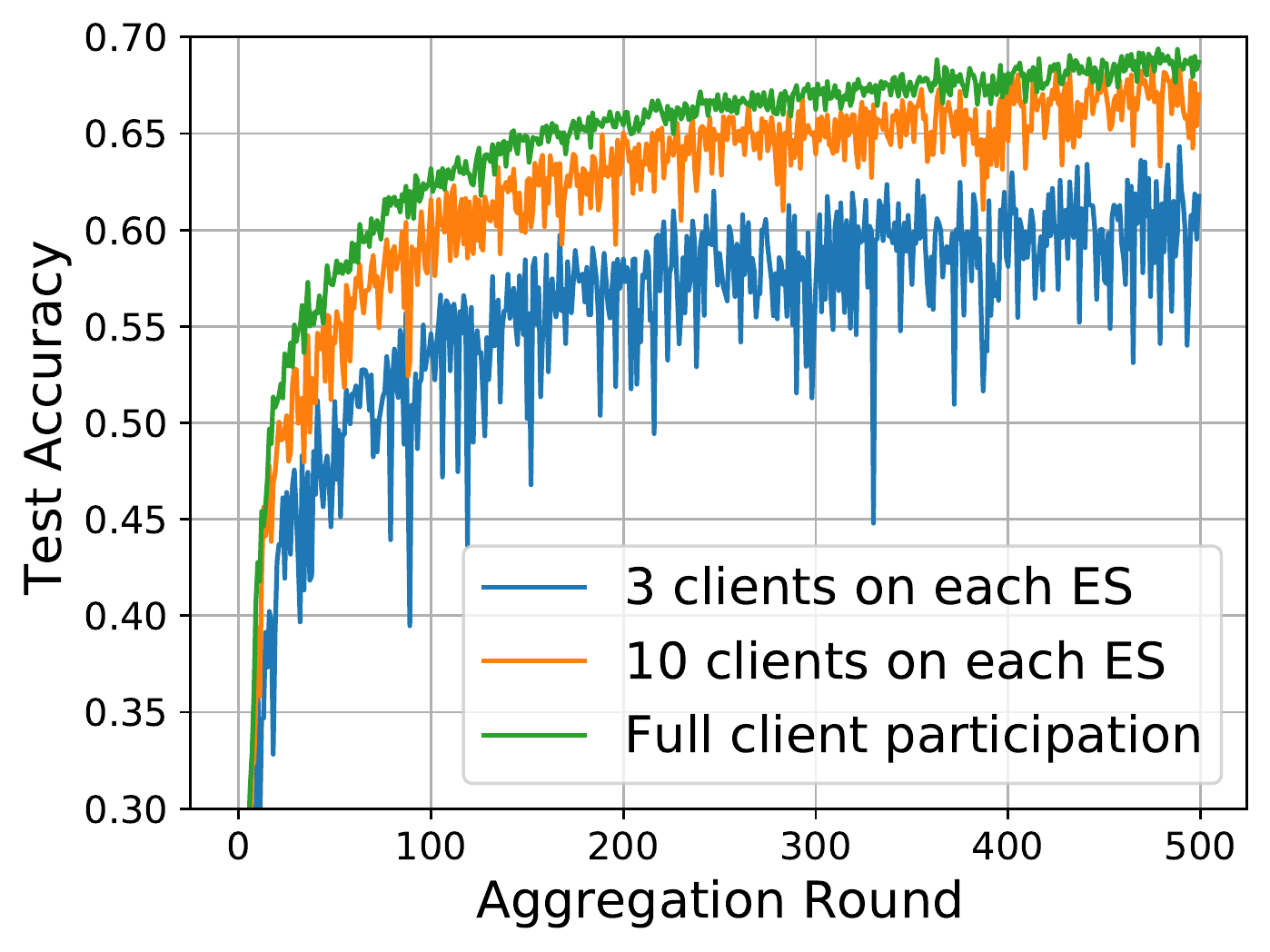}
      \subcaption{CIFAR-10 dataset under CNN.}
      \label{Fig:CIFARdiff}
      \end{minipage}
\caption{HFL training performance with different number of participating clients in each edge aggregation round.}
\label{Fig:training}
\end{figure}

\subsection{Utility Function of Client Selection of HFL}\label{SubSec:WNFL}
Some existing HFL studies \cite{wang2020local, liu2021hierarchical, luo2020hfel} have demonstrated that the convergence speed depends on the number of participating clients in each edge aggregation round for both strongly convex and non-convex HFL (i.e., the more clients participated, the faster convergence speed). In order to support the theoretical results, we show the training performance on our simulated HFL network with $M=3$ and $N=50$ in Fig.~\ref{Fig:training}, and it is observed that more participating clients on ESs can improve the performance in both the strongly convex and non-convex HFL settings.

For now, we consider strongly strongly convex HFL, where the convergence speed is linearly dependent on the number of participating clients. The client selection policy for non-convex HFL will be developed in Section~\ref{Sec:Extension}. As in \cite{nishio2019client, tran2019federated, yang2021characterizing}, not all the selected clients in $\bm{s}^t_m$ may reach the EC stage (i.e., $\sum_{n \in \bm{s}^t_m}X^{t}_n \leq S^t_m$) due to straggler drop-out. To achieve a targeted convergence criteria, NO thus needs to run more FL rounds, thereby incurring a higher training cost. Therefore, it is necessary to develop an efficient client selection policy to improve the convergence speed for HFL, where more clients can participate in every round without dropping out. Let $\bm{X}^t_m = \{X_{n,m}^t\}_{\forall n\in \mathcal{N}_m^t}$, then the utility of the client selection decision on ES $m$ is defined as:
\begin{equation}\label{Eq:stoutility}
    \mu (\bm{s}^t_m; \bm{X}^t_m )=\sum\nolimits_{n\in\bm{s}^t_m}X_{n,m}^t ,
\end{equation}
Further, let $\bm{s}^t = \{\bm{s}_1^t, \bm{s}_2^t, \dots, \bm{s}_M^t\}$ denote the client selection decision of the overall system and $\bm{X}^t = \{\bm{X}_1^t , \bm{X}_2^t , \dots, \bm{X}_M^t \}$. Therefore, the utility function of the whole HFL network is defined as:
\begin{equation}
    \mu (\bm{s}^t ;\bm{X}^t )=\frac{1}{M}\sum\nolimits_{n \in \mathcal{M}}\sum\nolimits_{n\in\bm{s}^t_m}X_{n,m}^t
\end{equation}

\subsection{Client Selection Problem Formulation}
The client selection problem for NO is a sequential decision-making problem. The goal of NO is to make selection decision $\bm{s}^t , \forall t$ to maximize the cumulative utility for a total of $T$ aggregation rounds. If an ES selects very few clients, its training performance may be degraded and the computation resources of computation resources of ESs may be wasted. To avoid bottleneck of HFL, we consider that NO equally divides the budget among the ESs (i.e., for each $m$, its budget is $B = \tilde{B}/M$). Assuming that NO knows a priori whether a selected client can return its model updates to the corresponding edge server in time, namely $\bm{X}^t, \forall t$, then the client selection problem is formulated: 
\begin{subequations}
\begin{align}
    \textbf{P1:}&~~~\max\nolimits_{\{\bm{s}^t \}_{t=1}^T}~~~\sum\nolimits_{t=1}^{T}\mu(\bm{s}^t ; \bm{X}^t)\label{Eq:cumulativeutility}\\
    &~~~\text{s.t.}~~~~\sum\nolimits_{n\in\bm{s}_m^t} c_n (y^t_n ) \leq B, ~~\forall m \in \mathcal{M} \label{Eq:budget}\\
    & ~~~~~~~~~~\bm{s}^t_m \subseteq \mathcal{N}_m^t , ~~\forall m, t \label{Eq:matroid}\\
    & ~~~~~~~~~~\bm{s}^t_m \cap \bm{s}^t_{m'} = \emptyset, ~~m,m'\in \mathcal{M}, \forall t.\label{Eq:empty}
\end{align}
\end{subequations}

The following challenges should be addressed to solve the client selection problem in HFL networks: (i) For maximizing the expected training utility of HFL, it is necessary to precisely estimate the selected clients in each edge aggregation round. In addition, since NO does not have enough experience to determine the selected clients at the first several rounds (i.e., cold start), collecting the historical data for estimation is important for this policy. (ii) With the successful participated clients estimation, how to optimize the selection decision on each ES under the limited budget should be carefully considered, because high variance of number of participated clients on each ES degrades the training performance. Therefore, we equally separate the total budget to each ES (constraint \eqref{Eq:budget}). (iii) Due to the movement of each client, the available connecting ESs can be considered as time-varying (constraint \eqref{Eq:matroid}), which brings more difficulties to make an efficient client selection decision. Note that constraint \eqref{Eq:empty} can guarantee that each client only can be selected to communicate at most one ES. (iv) Since the selection decisions are based on the estimated participated clients $\hat{\bm{X}}^t$, the accuracy of participated clients estimation will directly influence the training utility of NO. The following section will propose an policy based on the Multi-Armed Bandits (MAB) to address the mentioned challenges.

\section{Context-Aware Online Client Selection policy for Strongly Convex HFL}\label{Sec:COCS}
In this section, we formulate our client selection problem of HFL as a Contextual Combinatorial Multi-Armed Bandit (CC-MAB). The combinatorial property is because NO pays computation resources from multiple clients for maximizing the training utility. The contextual property is because NO leverages contexts associated with clients to infer their participated probabilities. In this paper, whether successfully participating in the corresponding ES depends on many side factors, which are collaboratively referred to as \textit{context}. We use contextual information to help infer the number of participated clients.

In CC-MAB, NO observes the context of clients at the beginning of each edge aggregation round before making the client selection decision. Recall that the participated probability of a client-ES pair depends on $r_{\text{DT}}^t , y_n^t$ and $r_{\text{UT}}^t$ in Eq.~\eqref{Eq:time}. Although it is difficult, if not impossible, for ESs to know the channel state of clients, the ESs can easily capture their own channel state to connected clients \cite{avin2009sinr, choi2020throughput}. Based on the channel state of DT and bandwidth $b_n^t$, ESs can compute the DT rate $r_{\text{DT}}^t$ . In addition, since the movement speed of clients is much slower than the transmission speed of wireless signals, while NO cannot know the UT rate $r_{\text{UT}}^t$, it is not difficult to be inferred by $r_{\text{DT}}^t$ (suppose that clients do not locate in the same area in each edge aggregation round). Note that computation resources $y_n^t$ of any client $n$ are revealed at the beginning of each round. Therefore, we set $r_{\text{DT}}^t$ and $y_n^t$ as \textit{context} and use these information to help infer the participated probabilities. Let $\phi_{n,m}^t \in \Phi$ denote the context of client-ES pair $(n,m)$ in edge aggregation round $t$. Without loss of generality, we normalize $\phi_{n,m}^t$ in a bounded space $\Phi = [0,1]^2$ using min-max feature scaling. Let $\bm{\phi}= \{\phi_{n,m}^t \}_{\forall m,n \in \mathcal{N}_m^t}$ denote the context of all clients on ESs. The context of all clients on ESs are collected in $\bm{\phi}^t = \{\phi_{n,m}^t\}_{\forall m , n\in \mathcal{N}_m^t}$. Whether successful participation of client $n$ on ES $m$ is a random variable parameterized by the context $\phi_{n,m}^t$. We slightly simplify the notation of selected clients and define the context-aware $X_{n,m}^t (\phi_{n,m}^t)$. Specifically, $X_{n,m}$ is a mapping function for each client-ES pair $(n,m)$, since the training time of clients is usually location-dependent, e.g., the distance between the client and ES, the communication environment, and other processing tasks on a client. We further define $p_{n,m} (\phi_{n,m}^t )\triangleq \mathbb{E}[X_{n,m} (\phi_{n,m}^t )]$ as the expected value (i.e., the participated probability $X_{n,m}^t \sim \text{Bernoulli}(p_{n,m}^t )$) of $X_{n,m}(\phi_{n,m}^t)$.

\subsection{Oracle Solution and Regret}\label{SubSec:oracle}
Similar to the existing CC-MAB studies \cite{chen2018contextual, chen2019budget, chen2020learning}, before providing our policy design, we first give an Oracle benchmark solution to the client selection problem of HFL by assuming that the NO knows the context-aware successful participated probability $p_{n,m}^t (\phi_{n,m}^t)$, $\forall m , n\in\mathcal{N}_m^t$. In this ideal setting, the utility function $\mu (\bm{s}^t ; \bm{p}^t )$ is perfectly known by NO, and thus we can get the optimal value of the client selection problem. The long-term selection problem $\textbf{P1}$ can be decomposed into $T$ independent subproblems in each round:
\begin{subequations}
\begin{align}
    \textbf{P2:}&~~~~~~~~~\max\nolimits_{\bm{s}^t}~~~\mu(\bm{s}^t ; \bm{p}^t)\label{Eq:singleutility}\\
    &~\text{s.t.}~~~~\sum\nolimits_{n\in\bm{s}_m^t} c_n (y^t_n ) \leq B, ~~\forall m \in \mathcal{M} \label{Eq:singlebudget}\\
    & ~~~~~~~~~~\bm{s}^t_m \subseteq \mathcal{N}_m^t , ~~\forall m, t \label{Eq:singlematroid}\\
    & ~~~~~~~~~~\bm{s}^t_m \cap \bm{s}^t_{m'} = \emptyset, ~~m,m'\in \mathcal{M}, \forall t.\label{Eq:singleempty}
\end{align}
\end{subequations}
\textbf{P2} is a combinatorial optimization problem with $M$ \textit{Knapsack} and a \textit{Matroid} constraints. The combinatorial property is because NO should choose a proper client selection decision to optimize participated probabilities on all ESs in order to achieve higher convergence speed. Knapsack constraints are from the constraint \eqref{Eq:singlebudget}, which bounds computation resources payment on each ES. 

To prove that \eqref{Eq:singlematroid} is a matroid constraint, we firstly state the definition of matroid. A matroid $\mathcal{E} = (\mathcal(X), \mathcal(I))$ is a system with independents sets, in which $\mathcal{X}$ is a finite set (named the ground set) and $\mathcal{I}$ represents the set of independent subsets of $\mathcal{X}$. It has the three following properties: (1) $\emptyset \in \mathcal{I}$ and $\mathcal{X}$ has at least one subset of $\mathcal{X}$; (2) For each $A \subset B \subset \mathcal{X}$, if $A \in \mathcal{I}$, then $B \in \mathcal{I}$; (3) If $A, B \in \mathcal{I}$, and $|A| < |B|$, then $\exists \in B \setminus A$ such that $A \{x\} \in \mathcal{I}$.

In the subproblem \textbf{P2}, let $\mathcal{X}=\cup_{m\in\mathcal{M}}\mathcal{N}_m^t$ denote the ground set of matroid $\mathcal{E}=(\mathcal{X},\mathcal{I})$, and $\mathcal{I}=\{I_1 ,I_2 ,\dots\}$ consists of subsets of $\mathcal{X}$ (i.e., $I_1 \subseteq \mathcal{X}$, $I_2 \subseteq \mathcal{X}$, $\dots$), where all $I \in \mathcal{I}$ includes at most one client from $\mathcal{N}_m^t$ for each $m \in \mathcal{M}$. We can write $I$ as $I = \cup_{m\in\mathcal{M}}\bm{s}_m^t$, s.t. $\bm{s}_m^t \in \mathcal{N}_m^t$, $\forall m$. In this paper, $\mathcal{I}$ is the set of all feasible client selection decisions. Therefore, it can be verified that Eq.~\eqref{Eq:singlematroid} is a matroid constraint \cite{chen2020learning, lee2004first}.

Based on our analysis, it is easy to observe that \textbf{P2} is NP-hard, and hence it can be solved by \textit{brute-force}, if the size of the HFL network is moderate. If the HFL network is too large, NO can use some commercial software to obtain the optimal solution, e.g., CPLEX \cite{cplex2009v12}. For simplicity, we define the optimal Oracle solution for each \textbf{P2} in edge aggregation round $t$ is $\bm{s}^{\text{opt},t}$. However, in practice, the prior knowledge of participated clients is infeasible, and thus the NO has to make a selection decision $\bm{s}^t$ based on the estimated participated clients $\hat{\bm{X}}^t$ in each edge aggregation round. Intuitively, NO should design an online client selection policy to choose $\bm{s}^t$ based on the estimation $\hat{\bm{X}}^t$. The performance of an online client selection policy is calculated by utility loss compared with the Oracle solution, called \textit{regret}. Suppose that we have a selection sequence $\{\bm{s}^1 ,\bm{s}^2,\dots, \bm{s}^T\}$ given by a policy, the expected regret is:
\begin{equation}\label{Eq:regret}
    \mathbb{E}[R(T)] = \sum\nolimits_{t=1}^T (\mathbb{E}[\mu(\bm{s}^{\text{opt},t};\bm{X}^t )]-\mathbb{E}[\mu(\bm{s}^t ;\bm{X}^t)]).
\end{equation}
The expectation is concerning with respect to the decisions made by the client selection decision policy and the participated clients over contexts.

\subsection{Context-aware Online Client Selection Policy}
Now, we will present our online client selection decision policy name Context-aware Online Client Selection (COCS). The COCS policy is designed based on CC-MAB. In edge aggregation round $t$, the process of COCS of NO is operated sequentially as follows: (i) NO observes the contexts of all client-ES pairs $\bm{\phi}^t = \{\phi_{n,m}^t\}_{n,m \in \mathcal{N}_m^t}$, $\phi_{n,m}^t \in \Phi$. (ii) NO determines its selection decision $\bm{s}^t$ based on the observed context information $\bm{\phi}^t$ in the current round $t$ and the knowledge learned from the previous $t-1$ rounds. (iii) The selection decision $\bm{s}^t$ is applied. If $s_n^t \neq 0$, $\forall n \in \bm{s}_m^t$, the clients located in the coverage area of ES $m$ can be selected by ES $m$ for training in round $t$. (iv) At the end of each edge aggregation round, the local model updates $\Delta_n^t$ from which clients are observed by all ESs, which is then used to update the estimated participated clients $\hat{X}_{n,m} (\phi_{n,m}^t)$ from the observed context $\phi_{n,m}^t$ of client-ES pair $(n,m)$.

The pseudocode of COCS policy is presented in Algorithm~\ref{Alg:COCS}. It has two parameters $K(t)$ and $h_T$ to be designed, where $K(t)$ is a deterministic and monotonically increasing function used to identify the under-explored context, and $h_T$ decides how we partition the context space. The COCS policy is stated as follows:

\begin{algorithm}[t!]
\caption{Context-aware Online Client Selection (COCS).}\label{Alg:COCS}
\begin{algorithmic}[1]
\Require $T$, $h_T$, $K(t)$.
\Ensure Create partition $\mathcal{L}_T$ on context space $\Phi$; set $C_{n,m}(l)=0$, $\mathcal{E}_{n,m} (l)=\emptyset$ $\hat{\beta}_{n,m} (l)=0$, $\forall n \in \mathcal{N}$ $\forall m \in \mathcal{M}$, $\forall l\in\mathcal{L}_T$.
\For {$t = 1, \dots, T$}
\State Observes the context of clients on ESs $\bm{\phi}^t$;
\State Determine $\mathcal{C}_n^{\text{ue},t}$ and $\mathcal{N}_n^{\text{ue},t}$, and estimate the participated clients $\hat{\bm{X}}^t$ based on the contexts $\bm{\phi}^t$;
\If {$\mathcal{N}_n^{\text{ue},t} \neq \emptyset$} \Comment{\textit{Exploration}}
\If {$\mathcal{N}_n^{\text{ue},t} \neq \emptyset, \forall n$}
\State Determine $\bm{s}^t$ by solving Eq.~\eqref{Eq:explorationi};
\Else
\State Get $\tilde{\bm{s}}^t$ by solving Eq.~\eqref{Eq:exploreiii};
\State Select the clients in $\mathcal{N}^{\text{ed},t}$ by solving Eq.~\eqref{Eq:mincase} and determine the client selection decision $\bm{s}^t$;
\EndIf
\Else \Comment{\textit{Exploitation}}
\State Determine the client selection decision $\S^t$ by solving the problem in Eq.~\eqref{Eq:exploitation};
\EndIf
\For {$n \in \mathcal{N}$} \Comment{\textit{Update}}
\State Identify each ES $m$ successfully receives the client $n$ before $\tau^{\text{dead}}$ and context hypercube $l$ that belongs to $\phi_{n,S_m}^t$;
\State Observe whether the selected client $n$ successfully participates on ES $m$ as $X_{n,m}$;
\State Update estimations: $\hat{p}_{n,m} (l) = \frac{\hat{p}_{n,m}(l) C_{n,m}(l) + X}{C_{n,m}(l)+1}$;
\State Update counters: $C_{n,m}(l) = C_{n,m}(l)+1$;
\EndFor
\EndFor
\end{algorithmic}
\end{algorithm}

\textbf{Initialization Phase:} Given parameter $h_T$, the proposed policy first creates a partition denoted by $\mathcal{L}_T$ for the context space $\Phi = [0,1]^2$, which splits $\Phi$ into $(h_T )^2$ sets. Each set is a $2$-dimensional hypercube with size $\frac{1}{h_T} \times \cdots \times \frac{1}{h_T}$. Note that $h_T$ is an important input parameter to guarantee policy performance. For each hypercube $l \in \mathcal{L}_T$, the NO keeps a counter $C^t_{n,m} (l)$ for each client $n \in \mathcal{N}$ and each ES $m \in \mathcal{M}$. For the tuple $(n,m,l)$ of a counter $C_{n,m}^t (l)$, we define a selection event $V_{n,m,l}$ that represents a selection decision satisfying the three following conditions: 1) the client $n \in \mathcal{N}_m^t$ is selected to an ES $m$; 2) the ES $m$ successfully receives the client $n$ before $\tau_{\text{dead}}$ (i.e., $\tau_{n}^t \leq \tau_{\text{dead}}$); 3) the context of client-ES pair $(n,m)$ belongs to $l$ (i.e., $\phi_{n,m}^t \in l$). The counter $C_{n,m}^t (l)$ stores the number of times that the event $V_{n,m,l}$ occurs until edge aggregation round $t$. Each ES $m$ also saves an experience $\mathcal{E}_{n,m}^t (l)$ for each client $n$ and each hypercube $l$, which contains the observed participation indicators when a selection event $V_{n,m,l}$ occurs. Based on the observed participation indicators in $\mathcal{E}_{n,m}^t (l)$, the estimated participated probability for a selection event $V_{n,m,l}$ is computed by:
\begin{equation}
    \hat{p}_{n,m}^t (l) = \frac{1}{C_{n,m}^t (l)}\sum\nolimits_{X \in \mathcal{E}_{n,m}^t (l)}X .
\end{equation}

In each edge aggregation round $t$, the COCS policy has the following phases:

\textbf{Hypercube Identification Phase:} If the local model updates $\Delta_n^t$ of client $n \in \mathcal{N}_m^t$ can be successfully received by an ES $m$ in edge aggregation round $t$, we obtain that $l_{n,m}^t$ is the hypercube for the context $\phi_{n,m}^t$, the estimated participated probability of client $n$ on ES $m$ is $\hat{X}_{n,m}^t = \hat{p}_{n,m}^t (l_{n,m}^t )$. Let $\hat{\bm{X}}^t = \{\hat{X}_{n,m}^t \}_{\forall m, n \in \mathcal{N}_m^t}$ denote the collection of all the estimated participated probabilities. For making a client selection decision, COCS policy needs to check whether these hypercubes have been explored sufficiently in order to ensure the enough accuracy of the estimated participated probability for each client-ES pair $(n,m)$. Therefore, we define \textit{under-explored} hypercubes $\mathcal{L}_m^{\text{ue}}(\bm{\phi}^t )$ for the ES $m$ in edge aggregation round $t$ as follows:
\begin{equation}\label{Eq:underexplored}
    \mathcal{L}^{\text{ue},t}_m \triangleq\left\{ l\in \mathcal{L}_T \bigg|
    \begin{aligned}
    & ~\exists \phi_{n,m}^t \in \bm{\phi}^t , \phi_{n,m}^t \in l , \tau_{n}^t \leq \tau_{\text{dead}}\\
    & ~\text{and}~ C_{n,m}^t (l) \leq K(t)
    \end{aligned}\right\}.
\end{equation}
Also, let $\mathcal{N}^{\text{ue},t}_m (\bm{\phi}^t )\triangleq \{n\in \mathcal{N}_m^t | l_{n,m}^t \in \mathcal{L}_m^{\text{ue},t}(\bm{\phi}^t )\}$ denote the collection of the under-explored client $n$ for each ES $m$. The challenge of COCS policy is how to decide the current estimated participated clients are accurate enough to guide the client selection decision in each edge aggregation round, which is referred as \textit{exploitation} or more training results need to be collected for a certain hypercube, which is referred to as \textit{exploration}. COCS policy aims to balance the exploration and exploitation phases in order to maximize the utility of NO up to a finite round $T$. Based on the $\mathcal{N}_m^{\text{ue},t}(\bm{\phi})^t $, COCS can identify that then either enters an exploration phase or an exploitation phase.

\textbf{Exploration Phase:} Firstly, let $\mathcal{N}_m^{\text{ue},t}(\bm{\phi}^t ) \triangleq \{n\in\mathcal{N}_m^t | \mathcal{N}_m^{\text{ue},t} \neq \emptyset \}$ denote an ES $m$ has under-explored clients, and $\mathcal{N}_m^{\text{ed},t}(\bm{\phi}^t )\triangleq \mathcal{N} \setminus \mathcal{N}^{\text{ue},t}(\bm{\phi}^t)$ denote the ES $m$ does not have under-explored clients. If the ES $m$ has a non-empty $\mathcal{N}_m^{\text{ue},t}$, then COCS enters the exploration phase. We may have two following cases in exploration phase: 

(i) All the clients have under-explored ESs. Intuitively, NO hopes to receive more local training updates $\Delta_n^t$. Therefore, COCS policy aims to select as many clients that have under-explored ESs sequentially solved by the following optimization:
\begin{equation}\label{Eq:explorationi}
    \max\nolimits_{\bm{s}^t}~|\bm{s}^t |~~~~~\text{s.t.}~ \text{\eqref{Eq:singlebudget}, \eqref{Eq:singlematroid}, \eqref{Eq:singleempty}},
\end{equation}
where $|\bm{s}^t |$ is the size of the collection $\bm{s}^t = \{\bm{s}_1^t ,\bm{s}_2^t ,\dots, \bm{s}_M^t \}$.

(ii) Part of ESs have under-explored clients $\exists \mathcal{N}_m^{\text{ue},t} \neq \emptyset$. We divide this case into two stages: NO firstly selects ESs that have under-explored clients $m \in \mathcal{N}_m^{\text{ue},t}$ by solving the following optimization:
\begin{subequations}\label{Eq:exploreiii}
\begin{align}
    &\max_{\tilde{\bm{s}}^t}~~~~|\tilde{\bm{s}}^t| \\
    \text{s.t.}~~~~& \sum_{n \in \tilde{\bm{s}}_m^t}c_n (y_n^t ) \leq B , ~~~~\forall n\in \mathcal{N}_m^{\text{ue},t}, \forall m \in \mathcal{M}\label{Eq:exploreiii2}\\
    & s^t_m \in \mathcal{N}^t_m \cup \{null\}, ~~~~~\forall n\in\mathcal{N}_m^{\text{ue},t} \label{Eq:exploreiii3}\\
    & \tilde{\bm{s}}_m^t \cap \tilde{\bm{s}}_{m'}^t = \emptyset , ~~m,m' \in \mathcal{M}, \forall t .
\end{align}
\end{subequations}
where $\tilde{\bm{s}}^t$ is client selection decision on ES $m$ that has under-explored clients and $|\tilde{\bm{s}}^t |$ is the size of the collection $\tilde{\bm{s}}^t = \{\tilde{\bm{s}}^t_1 ,\tilde{\bm{s}}^t_2 ,\dots, \tilde{\bm{s}}^t_M \}$. Secondly, ESs aim to select the explored clients $\forall n \in \mathcal{N}_m^{\text{ed},t}$. Here, we assume that there exists ESs that $B - \sum_{n\in \tilde{\bm{s}}_m^t}c_n (y_n^t ) \geq c^{\text{min}, t}, m\in \mathcal{M}$, where $c^{\text{min}, t} = \min_{n \in \mathcal{N}_m^{\text{ed},t}}c_n (y_n^{t}), \forall m$. Therefore, ESs can select the clients $n \in \mathcal{N}_m^{\text{ed},t}$ with the following constraint:
\begin{equation}\label{Eq:EDselection}
    \sum_{n \in \bm{s}_m^t \setminus \tilde{\bm{s}}_m^t}c_n (y_n^t ) \leq B - \sum_{n\in\tilde{\bm{s}}_m^t} c_n (y_n^{t}), ~\forall n \in \mathcal{N}_m^{\text{ed},t}.
\end{equation}
If not, NO does not need to select clients in $\mathcal{N}_m^{\text{ed},t}$ due to no budget left. Under this condition, the client selection decisions are jointly optimized the following optimization:
\begin{subequations}\label{Eq:mincase}
\begin{align}
    &\max_{\bm{s}^t}~\mu(\bm{s}^t ;\hat{\bm{X}}^t)\label{Eq:min}\\
    \text{s.t.}~ &B - \sum_{n\in \tilde{\bm{s}}_m^t}c_n (y_n^t ) \geq c^{\text{min}, t},~~ m\in \mathcal{M}, \label{Eq:restbudget}\\
    &\text{\eqref{Eq:exploreiii2}, \eqref{Eq:exploreiii3}, \eqref{Eq:EDselection}}.
\end{align}
\end{subequations}

\textbf{Exploitation Phase:} If the set of under-explored clients is empty (i.e., $\mathcal{N}_m^{\text{ue},t} = \emptyset , \forall m$), then COCS policy enters the exploitation phase. The optimal client selection decision $\bm{s}^t$ is derived by solving \textbf{P2} from the current estimated participated clients $\hat{\bm{X}}^t$:
\begin{equation}\label{Eq:exploitation}
    \max_{\bm{s}^t}~~\mu(\bm{s}^t ;\hat{\bm{X}}^t)~~~\text{s.t.}~ \text{\eqref{Eq:singlebudget}, \eqref{Eq:singlematroid}}.
\end{equation}

\textbf{Update Phase:} After selecting the client-ES pair in each round $t$, the proposed COCS policy observes whether the local model updates of selected clients can be received before the deadline $\tau^{\text{dead}}$; then, it updates $\hat{p}_{n,m}^t (l)$ and $C^t_{n,m} (l)$ of each hypercube $l\in\mathcal{L}_T$.

\subsection{Performance Analysis}
To present an upper performance bound of the COCS policy in terms of \textit{regret}, we make the following assumption that the participated clients are similar are similar when their contexts are similar. This natural assumption is formalized by the following H\"{o}lder condition \cite{chen2018contextual, chen2019budget, chen2020learning}, which is defined as follows:

\begin{assumption}\label{Ass:holder}
(H\"{o}lder Condition). If a real function $f$ on $D$-dimensional Euclidean space satisfies H\"{o}del condition, there exists $L>0$, $\alpha >0$ such that for any $\phi, \phi' \in \Phi$, it holds that $|f_n (\phi)-f_n(\phi')|\leq L\|\phi-\phi'\|^\alpha$ for an arbitrary client $n\in\mathcal{N}$, where $\|\cdot \|$ is the Euclidean norm.
\end{assumption}

By providing the design of the input parameters $K(t)$ and $h_T$, we show that COCS policy achieves a sublinear $R(T)=O(T^\gamma)$ with $\gamma <1$, which guarantees that COCS has an asymptotically optimal performance. This means that the online client selection decision made by COCS policy converges to the Oracle solution. Because any edge aggregation round is either in the exploration or exploitation phase, the regret can be divided into two parts $R(T) = R_{\text{explore}}(T) + R_{\text{exploit}}(T)$, where $R_{\text{explore}}(T)$ and $R_{\text{exploit}}(T)$ are the regrets due to exploration and exploitation phases, respectively. The total regret bound is achieved by separately bounding these two parts. Therefore, we present the following two lemmas for bounding exploration and exploitation regrets.

\begin{lemma}\label{lemma:boundexplore}
(Bound of $\mathbb{E}[R_{\text{explore}}(T)]$.) Given the input parameters $K(t) = t^z \log(t)$ and $h_T = \lceil T^\gamma \rceil$, where $0 < z<1$ and $0 < \gamma <\frac{1}{2}$, the regret $\mathbb{E}[E_{\text{explore}}(T)]$ is bounded by:
\begin{equation}
    \mathbb{E}[E_{\text{explore}}(T)] \leq \frac{4N^2 MB}{c^{\text{min}}}( T^{z+2\gamma} \log(T) + T^{2\gamma}),\nonumber
\end{equation}
where $c^{\text{min}}=\min_{y^t_n , \forall n,t}c_n (y^t_n )$.
\end{lemma}

\noindent\textit{Proof.} See Appendix~A in the supplemental file. ~\hfill$\Box$

Lemma~\ref{lemma:boundexplore} shows that the order of $R_{\text{explore}}(T)$ is determined by the control function $K(T)$ and the number of hypercubes $(h_T)^D$ in partition $\mathcal{L}_T$.

\begin{lemma}\label{lemma:boundexploit}
(Bound of $\mathbb{E}[R_{\text{exploit}}(T)]$.) Given $K(t) = t^z \log(t)$ and $h_T = \lceil T^\gamma \rceil$, where $0 <z<1$ and $0< \gamma <\frac{1}{2}$, if the H\"{o}lder condition holds true and the additional condition $2H(t) + \frac{2NMB}{c^{\text{min}}}L2^{\frac{\alpha}{2}}h_T^{\alpha}\leq At^\theta$ is satisfied with $H(t)>\frac{NMB}{c^{\text{min}}}t^{-\frac{z}{2}}$, $A>0$, $\theta <0$, for all $t$, then $\mathbb{E}[R_{\text{exploit}}(T)]$ is bounded by:
\begin{equation}
\begin{split}
    \mathbb{E}[R_{\text{exploit}}(T)] &\leq \frac{NMB}{c^{\text{min}}}\bigg(\sum_{k=1}^{B/c^{\text{min}}}\begin{pmatrix}
N\\
k
\end{pmatrix}\bigg)\frac{\pi^2}{3}\\
    & + \frac{3NMB}{c^{\text{min}}}L2^{\frac{\alpha}{2}} T^{1-\gamma \alpha} + \frac{A}{1+\theta}T^{1+\theta}, \nonumber
\end{split}
\end{equation}
where $c^{\text{min}}=\min_{y^t_n , \forall n,t}c_n (y^t_n )$.
\end{lemma}

\noindent\textit{Proof.} See Appendix~B in the supplemental file. ~\hfill$\Box$

Lemma~\ref{lemma:boundexploit} indicates that the regret of exploitation $\mathbb{E}[R_{\text{exploitation}}(T)]$ depends on the choice of $z$ and $\gamma$ with an additional condition being satisfied. Based on the above two Lemmas, we will have the following Theorem for the upper bound of the regret $\mathbb{E}[R(T)]$.

\begin{theorem}\label{theorem:RT}
(Bound of $\mathbb{E}[R(T)]$.) Given the input parameters $K(t) = t^z \log(t)$ and $h_T = \lceil T^\gamma \rceil$, where $0 <z<1$ and $0< \gamma <\frac{1}{2}$, if the H\"{o}lder condition holds true and the additional condition $2H(t) + \frac{2NMB}{c^{\text{min}}}L 2^{\frac{\alpha}{2}}h_T^{\alpha}\leq At^\theta$ is satisfied with $H(t)>\frac{NMB}{c^{\text{min}}}t^{-\frac{z}{2}}$, $A>0$, $\theta <0$, for all $t$, then the regret $\mathbb{E}[R(T)]$ can be bounded by:
\begin{equation}
\begin{split}
    \mathbb{E}[R(T)] &\leq \frac{4N^2 MB}{c^{\text{min}}}( T^{z+2\gamma} \log(T) + T^{2\gamma}) \\
    & + \frac{NMB}{c^{\text{min}}}\bigg(\sum_{k=1}^{B/c^{\text{min}}}\begin{pmatrix}
N\\
k
\end{pmatrix}\bigg)\frac{\pi^2}{3}\\
    & + \frac{3NMB}{c^{\text{min}}}L2^{\frac{\alpha}{2}} T^{1-\gamma \alpha} + \frac{A}{1+\theta}T^{1+\theta}, \nonumber
\end{split}
\end{equation}
where $c^{\text{min}}=\min_{y^t_n , \forall n,t}c_n (y^t_n )$.
\end{theorem}

\noindent\textit{Proof.} See Appendix~C in the supplemental file. ~\hfill$\Box$

The regret upper bound in Theorem~\ref{theorem:RT} is given with properly choosing input parameters $K(t)$ and $h_t$. However, the values of $z, \gamma, A$ and $\theta$ are not deterministic. Next, we will show that the regret upper bound of $\mathbb{E}[R(T)]$ in these parameters design.

\begin{theorem}\label{theorem:upperbound}
(Regret upper bound). If we select $z = \frac{2\alpha}{3\alpha +2} \in (0,1)$, $\gamma = \frac{z}{2\alpha}$, $\theta = -\frac{z}{2}$, $A = \frac{2NMB}{c^{\text{min}}}t^{-\frac{z}{2}} + \frac{2NMB}{c^{\text{min}}}L2^{\frac{\alpha}{2}}$, and COCS algorithm runs with these parameters, the regret $\mathbb{E}[R(T)]$ can be bounded by:
\begin{equation}
\begin{split}
    \mathbb{E}[R(T)] & \leq \frac{4N^2 MB}{c^{\text{min}}}(\log(T)T^{\frac{2\alpha + 2}{3\alpha +2}} + T^{\frac{2}{3\alpha + 2}})\\
    & + \frac{NMB}{c^{\text{min}}}\bigg(\sum_{k=1}^{B/c^{\text{min}}}\begin{pmatrix}
N\\
k
\end{pmatrix}\bigg)\frac{\pi^2}{3}\\
& + \bigg(3L2^{\frac{\alpha}{2}} + \frac{2+L2^{\frac{\alpha}{2}}}{(2\alpha + 2)/(3\alpha + 2)}\bigg)\frac{NMB}{c^{\text{min}}}T^{\frac{2\alpha + 2}{3\alpha +2}}, \nonumber
\end{split}
\end{equation}
where $c^{\text{min}}=\min_{y^t_n , \forall n,t}c_n (y^t_n )$. The dominant order of the regret $\mathbb{E}[R(T)]$ is $O\left(\frac{4N^2 MB}{c^{\text{min}}}T^{\frac{2\alpha +2}{3\alpha +2}}\log(T)\right)$.
\end{theorem}

\noindent\textit{Proof.} See Appendix~C in the supplemental file. ~\hfill$\Box$

The dominant order of regret upper bound, which indicates the COCS policy in Theorem~\ref{theorem:upperbound} is sublinear. In addition, the regret bound is valid for any total rounds $T$, and it can be used to characterize the convergence speed of HFL. 

\subsection{Complexity Analysis} 
The space complexity of COCS policy is determined by the number of counters $C^t_{n,m} (l)$ and experiences $\mathcal{E}^t_{n,m} (l)$ maintained for hypercubes. Because the counter is an integer for each hypercube, the space complexity is determined by the number of hypercubes. The experience $\mathcal{E}^t_{n,m} (l)$ is a set of observed successfully participating clients records up to round $t$, which requires a higher memory. However, it is unnecessary to store all historical records, since most estimators can be updated recursively. Therefore, the NO only needs to keep the current participated clients estimation for a hypercube. If COCS is run with the parameters in Theorem~\ref{theorem:upperbound}, the number of hypercubes is $(h_T)^2 =\lceil T^{\frac{1}{3\alpha +2}} \rceil^2 $, and thus the required space is sublinear in total rounds $T$. This means that when $T\to \infty$, COCS will require infinite memory. In the practical implementations, NO only needs to keep the counters and experiences of hypercubes to which at least one of the observed contexts occurs. Therefore, the practical space requirement of some counters and experiences is much smaller than the theoretical requirement.

\section{COCS for Non-convex HFL}\label{Sec:Extension}
In this section, we will discuss the solutions for the client selection problems for non-convex HFL, e.g., neural network. The convergence speed weakly depends on the number of participated clients in traditional FL \cite{karimireddy2020scaffold, yang2021achieving}, i.e., $O(\frac{1}{\sqrt{|\bm{s}^t |}T})$. In HFL, the convergence speed has similar performance $O(\frac{1}{\sqrt{\frac{1}{M}\sum_{m\in\mathcal{M}}|\bm{s}_m^t |T}})$ \cite{liu2021hierarchical, wang2020local}. Based on the utility function of strongly convex HFL in Eq.~\eqref{Eq:stoutility}, we define the non-convex HFL utility function as follows:
\begin{equation}\label{Eq:nonconvexutility}
    \mu_{\text{non}} (\bm{s}^t ; \bm{X}^t )=\sqrt{\frac{1}{M}\sum_{m\in\mathcal{M}}\sum_{n\in\bm{s}^t_m} X_{n,m}^t}.
\end{equation}
Therefore, the client selection problem of non-convex HFL in edge aggregation round $t$ is formulated as follows:
\begin{equation}\label{Eq:nonsingleutility}
    \textbf{P3:}~\max_{\bm{s}^t}~~\mu_{\text{non}}(\bm{s}^t ;\bm{X}^t )~~~\text{s.t.}~\text{\eqref{Eq:singlebudget}, \eqref{Eq:singlematroid}}.
\end{equation}
The problem \textbf{P3} is also a combinatorial optimization problem. While \textit{brute-force} search can always find the optimal solution, the complexity can be high due to the non-linear property in Eq.~\eqref{Eq:nonsingleutility}. In order to address this problem, we aim to design an efficient polynomial runtime approximation algorithm to solve \textbf{P3} in the next subsection. In addition, the performance guarantee of the approximation algorithm will be presented. 

\subsection{Approximated Oracle Solutions}
To solve the problem \textbf{P3}, We first show that \textbf{P3} is a monotone submodular maximization problem with $M$ knapsack and a matroid constraints. Below gives the definition of the monotone submodular maximization \cite{lee2004first}:
\begin{definition}\label{Def:submodular}
(Monotone Submodular Maximization.) A set function $F:2^{I}\to \mathcal{R}$ is monotone increasing if $\forall A \subseteq B \subseteq I$, $F(A)\leq F(B)$. In addition, the function $F(\cdot)$ is submodular if $\forall A\subseteq B \subseteq I$ and $e \in I\setminus B$, $F(A \cup \{e\})-F(A) \geq F(B \cup \{e\})-F(B)$.
\end{definition}
\begin{theorem}\label{The:submodular}
\textbf{P3} is a monotone submodular maximization with $M$ knapsack and a matroid constraints problem. 
\end{theorem}

\noindent\textit{Proof.} See Appendix~D in the supplemental file. ~\hfill$\Box$ 

To facilitate the solution of client selection for non-convex HFL in \textbf{P3}, approximation algorithms are efficiently obtained approximate solutions in polynomial runtime. Some existing studies are focusing on solving the submodular maximization with knapsack and matroid constraints \cite{badanidiyuru2014fast, chekuri2014submodular, mirzasoleiman2016fast}, and they proposed the approximation guarantee to the optimal solution. In this paper, we use the Fast Lazy Greedy (FLGreedy) algorithm in \cite{badanidiyuru2014fast} to achieve the approximated oracle solution with $\frac{1}{(1+\epsilon)(2+2M)}$-approximation guarantee, where $\epsilon$ is an error parameter of the algorithm. 

For coherence, we do not introduce the detailed FLGreedy algorithm here. The FLGreedy algorithm acquires the client selection decision of the client sequentially, which starts with the all-$null$ decisions. In each edge aggregation round, it selects a client to an ES that gives the largest incremental learning utility. Because each client can only be selected at most one ES and each iteration decides the client selection decision for one client, the algorithm terminates in at most $M$ iterations. Based on the results for submodular maximization with a knapsack and a matroid constraints in \cite{badanidiyuru2014fast}, the FLGreedy algorithm guarantees to yield a $\frac{1}{(1+\epsilon)(2+2M)}$-approximation for \textbf{P3}:
\begin{lemma}
In an arbitrary edge aggregation round $t$, let $\bm{s}^{*,t}$ be the client selection decision solved by FLGreedy algorithm and $\bm{s}^{\text{opt},t}$ be the optimal client selection decision for the problem \textbf{P3}, we will have $\mu(\bm{s}^{*, t}; \bm{X}^t ) \geq \frac{1}{(1+\epsilon)(2+2M)}\mu (\bm{s}^{\text{opt},t}; \bm{X}^t )$.
\end{lemma}

\noindent\textit{Proof.} The proof follows \cite{badanidiyuru2014fast} and hence is omitted. ~\hfill$\Box$

We use FLGreedy to approximate the optimal client selection decision with oracle information on participated clients. Note that the actual performance of the FLGreedy algorithm is usually much better than the $\frac{1}{(1+\epsilon)(2+2M)}$ approximation ratio in practice.

\subsection{Performance Analysis for Non-Convex HFL}
The regret in Eq.~\eqref{Eq:regret} is used when the optimal oracle solutions can be derivable. Because the FLGreedy algorithm can efficiently approximate the optimal oracle solution for \textbf{P3} instead of obtaining the optimal oracle solution. As such, we leverage the definition of $\delta$-regret, which is usually used in MAB based on approximation algorithms \cite{chen2017interactive}. For a $\delta$-approximation algorithm, i.e., the solution $\bm{s}^t$ solved by the approximation algorithm satisfies $\mu(\bm{s}^t ; \bm{X}^t) \geq \frac{1}{\delta}\mu(\bm{s}^{\text{opt}, t} ; \bm{X}^t )$, for problem \textbf{P3}, the $\delta$-regret is defined as follows:
\begin{equation}\label{Eq:deltaregret}
    R^\delta (T) = \sum_{t=1}^{T}\frac{1}{\delta}\mu(\bm{s}^{\text{opt},t}; \bm{X}^t ) - \sum_{t=1}^{T}\mu(\bm{s}^{t} ; \bm{X}^t ).
\end{equation}
Because the FLGreedy algorithm for the problem \textbf{P3} has an approximation ratio of $\frac{1}{(1+\epsilon)(2+2M)}$, COCS policy obtains $\delta = \frac{1}{(1+\epsilon)(2+2M)}$. The definition of $\delta$-regret essentially compares the utility of a policy with the lower bound of the approximated oracle solution.

Next, we aim to present the input parameters $K(t)$ and $h_T$ and propose a regret upper bound for COCS policy. The regret analysis is also proved based on the H\"{o}lder condition in Assumption~\ref{Ass:holder}. Given the input parameters $K(t)$ and $h_T$ in Theorem~\ref{The:appbound}, we achieve a sublinear regret upper bound of COCS policy for \textbf{P3} as follows:

\begin{theorem}\label{The:appbound}
($\delta$-Regret Upper Bound.) If $K(t) = t^{\frac{2\alpha +2}{3\alpha +2}}\log(t)$, $h_T = \lceil T^{\frac{1}{3\alpha +2}} \rceil$, H\"{o}lder condition holds true and a $\delta$-approximation is applied for optimization, then the dominate order of $\delta$-regret $\mathbb{E}[R^{\delta} (T)]$ is $O\left(\frac{4N^2 MB}{\delta c^{\text{min}}} T^{\frac{2\alpha +2}{3\alpha +2}}\log(T)\right)$.
\end{theorem}

\noindent\textit{Proof.} See Appendix~E in the supplemental file. \hfill~$\Box$

The regret upper bound given in Theorem~\ref{The:appbound} implies that COCS policy performs well enough if the subproblem in each edge aggregation round can only be derived approximately. A sublinear $\delta$-regret can be achieved based on the performance guarantee of $\delta$-approximation algorithms.

\section{Simulations}\label{Sec:Experiments}
In this section, we conduct simulations to evaluate the performance of our proposed COCS policy of HFL. Firstly, we introduce the simulation setup in this paper. Then, we move to evaluate the performance of strongly convex and non-convex HFL by different benchmarks respectively. The simulation results will show how our proposed COCS policy impacts the convergence and ML model performance.

\subsection{Setup}
\textbf{Datasets and Training models:} We set up the simulation with PyTorch and the computation is conducted by a high-performance workstation with 2 NVIDIA RTX 2080 GPUs. We have prepared two datasets for evaluating our proposed COCS algorithm. Specifically, MNIST dataset \cite{lecun1998gradient} under a logistic regression, which is widely used for strongly convex FL studies \cite{li2019convergence, yang2019parallel}. For the non-convex HFL, we use CIFAR-10 dataset \cite{krizhevsky2009learning} by adopting a CNN with two $5 \times 5$ convolution layers (each with $64$ channels), followed with $2\times 2$ max polling, two fully-connected layers with 384 and 192 units, and finally a softmax output layer. For each simulation, we distribute the dataset among $N=50$ clients in a general non-iid fashion such that each clients only contains samples of only two labels.

\begin{table}[t!]
\centering
\caption[HFL]{HFL Network Parameters (If the parameter has two values, the first one is on MNIST dataset and second is on CIFAR-10 dataset).}
\begin{tabular}{c|c}
\hline
\textbf{Parameter} & \textbf{Value} \\\hline
Number of clients, $N$              &     50                         \\ \hline
Number of ESs, $M$                  &     3                         \\ \hline
Size of local model updates, $s$     &   0.18Mbits, 18.7Mbits                 \\\hline
Computation workload of a client, $q^n$          &        2.41Mbytes, 28.3Mbytes                         \\\hline
Transmission power of a client, $P_n^t$          &            23dBm                     \\\hline
Deadline, $\tau^{\text{dead}}$          &      3sec, 20sec                     \\\hline
Pricing function, $b_n (f_n )$       &          $\mathcal{U} \sim [0.5, 2]$ per MHz           \\\hline
Budget on each ES $B$   & 3.5, 40           \\\hline
Dimension of context space, $D$     &           4                      \\\hline
The parameter $\alpha$ in H\"{o}lder condition                 &          1                       \\\hline
$h_T$ in COCS                                    &               5                  \\\hline
Local training epochs, $E$                                    &      2, 5                           \\\hline
Global aggregation, $T_{ES}$        &  5 \\\hline
Learning rate, $\eta$                                    &      0.005, 0.1                          \\\hline
\end{tabular}
\label{Tab:parameters}
\end{table}

\textbf{Contexts:} For the context generation, in each edge aggregation round, we assume the allocated bandwidth of all clients is sampling from a uniform distribution between $\mathcal{U} \sim [0.3, 1]$MHz for MNIST dataset and $\mathcal{U} \sim [2, 4]$MHz for CIFAR-10 dataset, since the data and model sizes are different for each dataset. Likewise, the available computation capacity of all clients is also sampling from $\mathcal{U} \sim [2, 4]$MHz for MNIST dataset and $\mathcal{U} \sim [8, 15]$MHz for CIFAR-10 dataset. The distance $d_{n,m}^t$ between client and ES is from $\mathcal{U} \sim [0, 2]$km.

\textbf{Parameters of HFL Networks:} In our simulated HFL network, the radius of each ES is 2km. Within the coverage area, there are several clients randomly distributed and communicated by the corresponding ES through a wireless channel in each edge aggregation round. In the edge aggregation round $t$, the downlink and uplink channel gain are decomposed of both small-scale fading and large-scale fading, where the small-scale fading is set as Rayleigh distribution with uniform variance and the large-scale fading are calculated by the path-loss with random shadowing $g_{\text{DT},n}^t = g_{\text{UT},n}^t = 37.6\log(d_{n,m}^t ) + 128.1$, where $d$ represent the distance of client-ES pair $(n,m)$. To clearly show the performance difference of each benchmark, we set HFL network parameters for the two datasets shown in Table~\ref{Tab:parameters}.

\subsection{Comparison Benchmarks}
We compare the COCS policy with the four following benchmarks:

\textbf{1) Oracle:} the Oracle algorithm knows precisely whether one client can be received by the corresponding ES before the deadline $\tau^{\text{dead}}$ with any observed context. In each aggregation round, it makes a client selection decision to maximize the utility in Eq.~\eqref{Eq:singleutility}: brute-force for the strongly convex HFL and GreedyLS for non-convex HFL.

\textbf{2) Combinatorial UCB (CUCB):} CUCB is designed based on a classical MAB policy UCB \cite{auer2002finite}. It develops combinations of client selection decisions on all ESs to enumerate NO's decision $\bm{s}$. CUCB runs UCB with feasible NO selection decisions $\bm{s}$ and learns the expected utility for each $\bm{s}^t$ in edge aggregation round. Since CUCB does not fit for the time-varying arm set, we set the static computation and transmission resource for client-ES pairs.

\textbf{3) LinUCB:} LinUCB \cite{li2010contextual} is a contextual variant of running CUCB. LinUCB also aims to learn the expected utility for client selection decision $\bm{s}$, which assumes that the utility of an arm is a linear function of client-ES pairs' contexts.

\textbf{4) Random:} The Random algorithm selects a client to an accessible ES randomly in each edge aggregation round under these two constraints.

\begin{figure}[t!]
  \centering
  \begin{minipage}{0.48\columnwidth}
      \includegraphics[draft=false,width=\linewidth, height=0.65\linewidth]{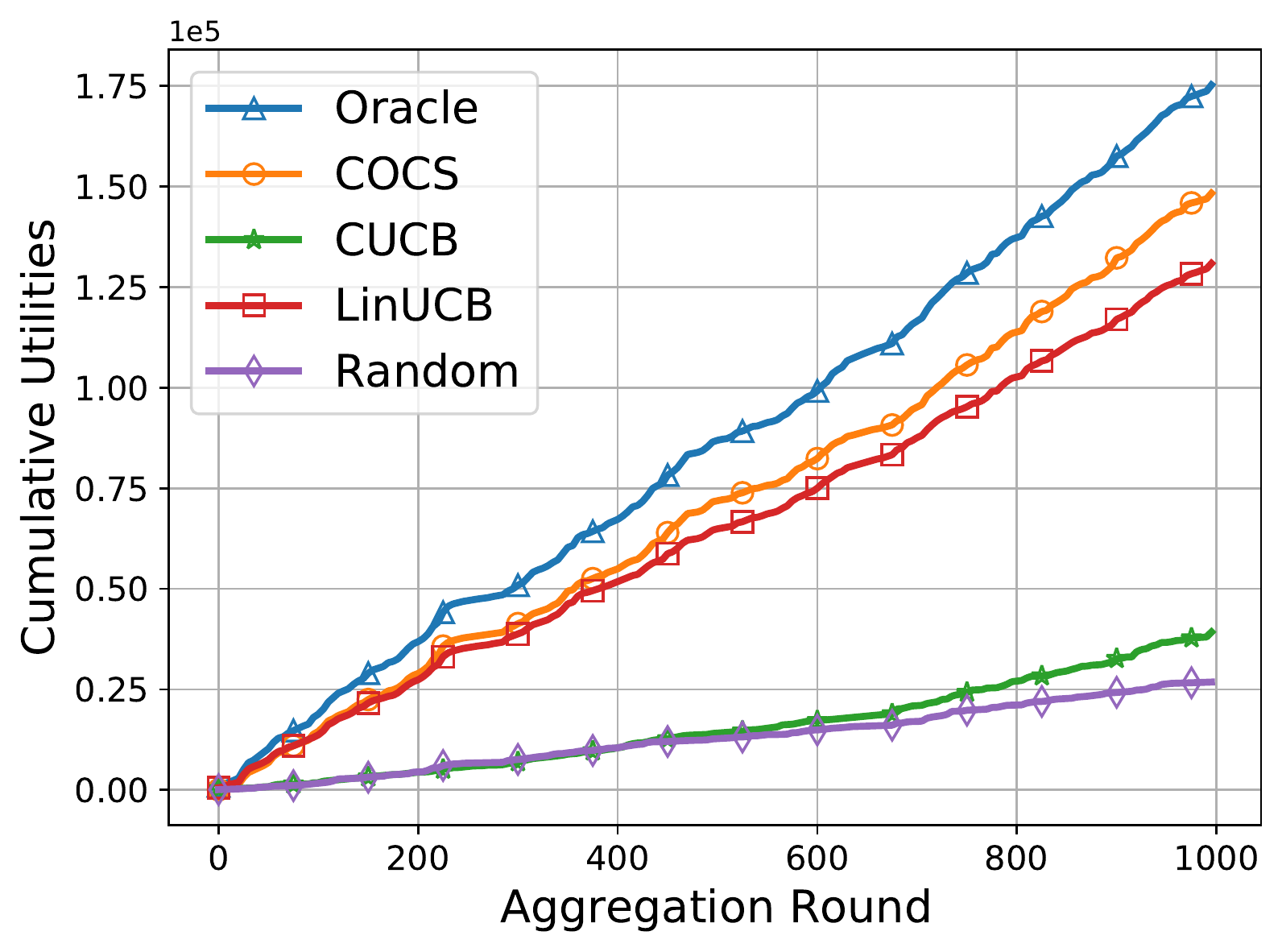}
      \subcaption{Cumulative Utilities.}
      \label{Fig:mnist_COCSutility}
    \end{minipage}
    \hfill
    \begin{minipage}{0.48\columnwidth}
      \includegraphics[draft=false,width=\linewidth, height=0.65\linewidth]{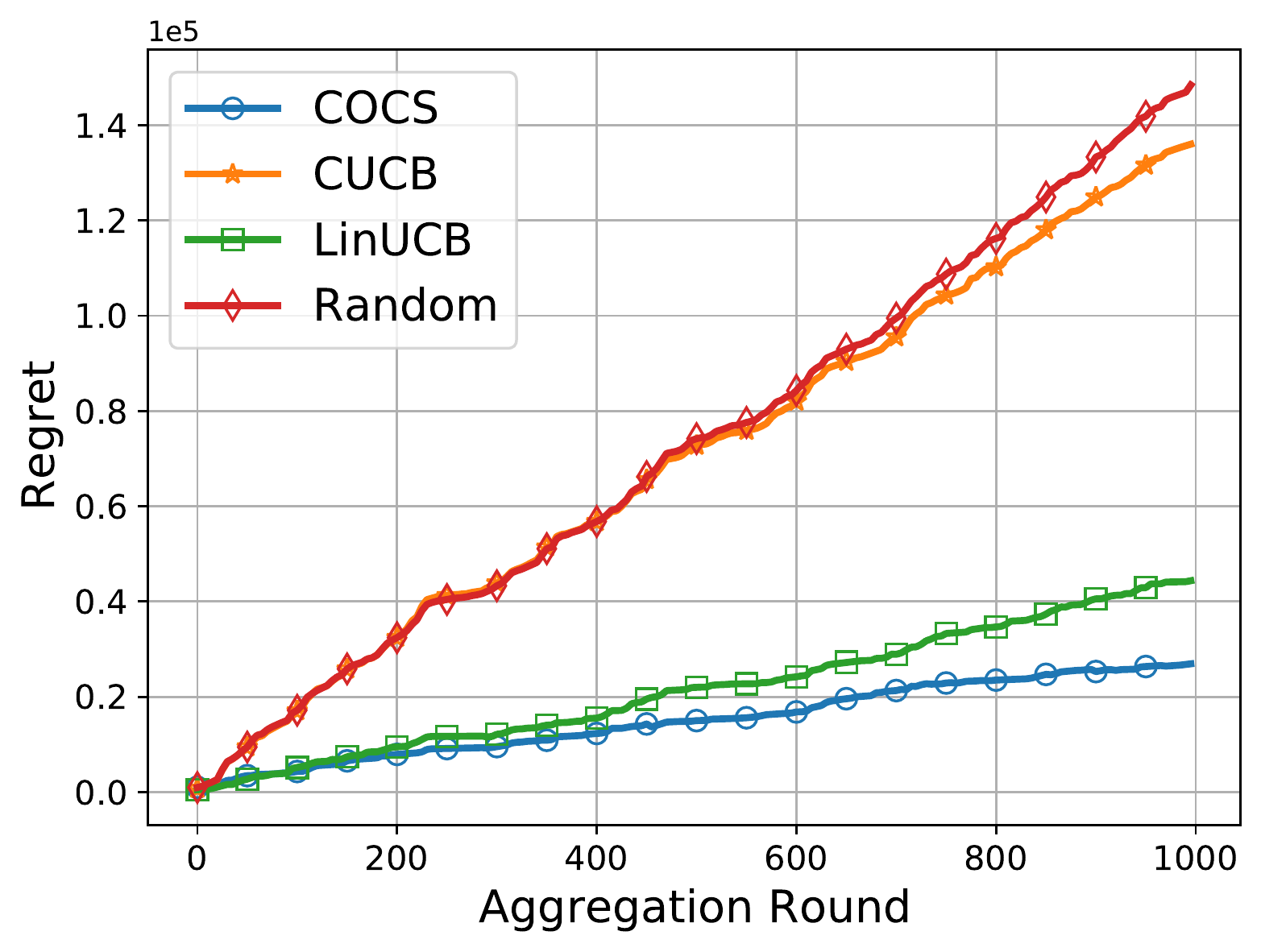}
      \subcaption{Regret.}
      \label{Fig:mnist_COCSregret}
      \end{minipage}
\caption{Comparisons on cumulative utilities under logistic regression on MNIST dataset.}
\end{figure}

\subsection{Performance Evaluation of Strongly Convex HFL}
\textbf{1) Comparison on cumulative utilities:} Fig.~2 shows the cumulative utilities and regret obtained by the COCS policy and the other 4 benchmarks during 1,000 edge aggregation rounds under logistic regression on the MNIST dataset. For the cumulative utilities in Fig.~\ref{Fig:mnist_COCSutility}, it is observed that Oracle achieves the highest cumulative utilities and provides an upper bound to the other benchmarks as expected. Among the others, COCS policy significantly outperforms the other benchmarks and has a closed cumulative utility performance to Oracle. The profit of the context of client-ES pairs can be shown by comparing the performance of context-aware policy (LinUCB) and context-unaware policies (CUCB and Random). Moreover, the results indicate that the cumulative utilities of CUCB are similar to the Random policy. The disadvantage of CUCB is from the following two reasons: (1) an arm of CUCB is a combination of the selection decisions of all client-ES pairs, and thus CUCB has a large number of arms. This means that CUCB is difficult to enter the exploitation phase. (2) CUCB fails to capture the connection between context and clients. The cumulative utilities of LinUCB are based on the context, for which a CUCB arm is not effective to produce a good result due to the large arm set. Fig.~\ref{Fig:mnist_COCSregret} notably depicts the regret generated by the 5 benchmarks. It is easy to observe that our proposed COCS policy incurs a sublinear regret.

\begin{figure*}[t!]
\begin{minipage}{0.32\columnwidth}
  \vspace*{\fill}
  \centering
  \includegraphics[draft=false,width=\linewidth, height=0.72\linewidth]{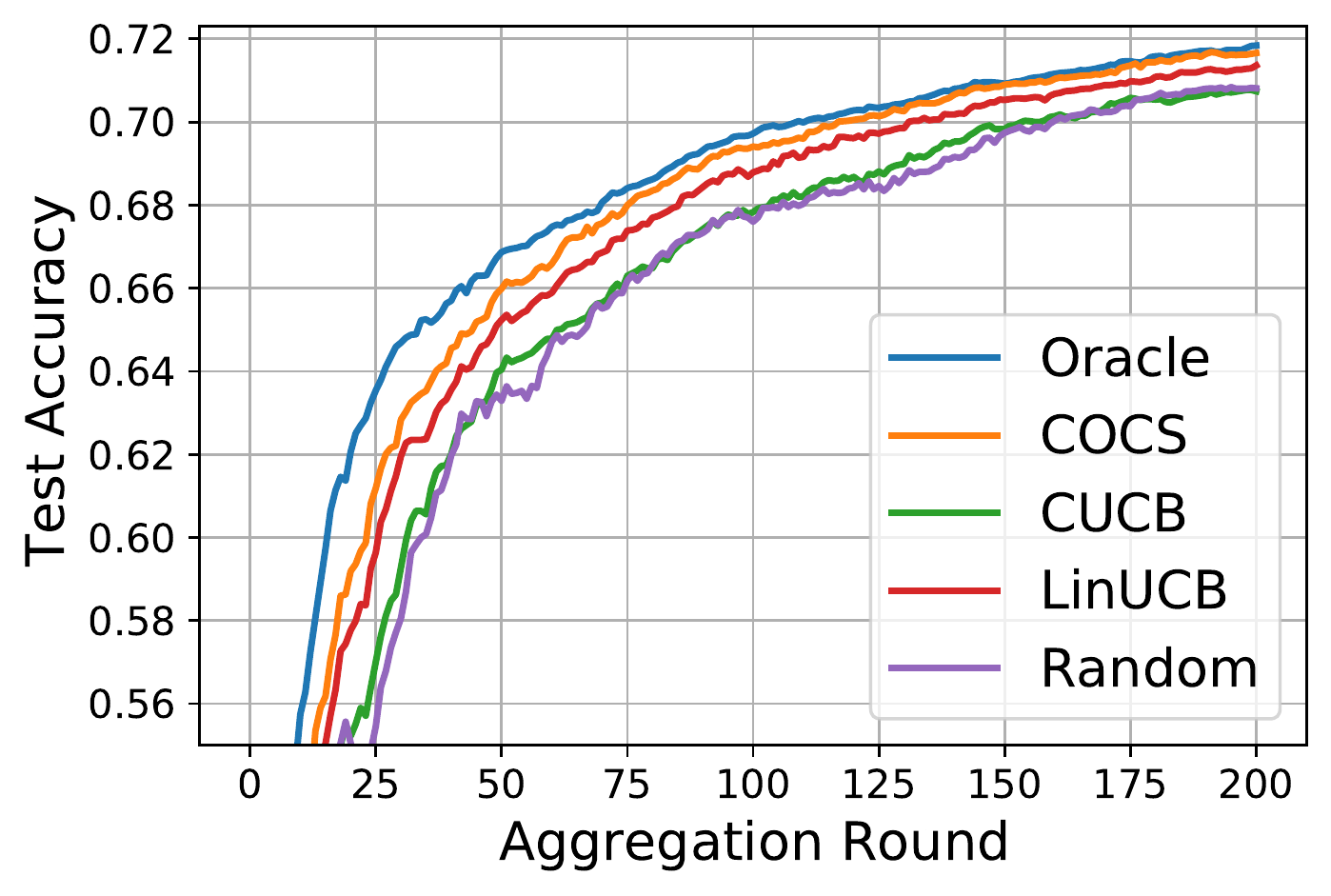}
  \subcaption{}
  \label{Fig:mnist_performance}\par\vfill
  \includegraphics[draft=false,width=\linewidth, height=0.72\linewidth]{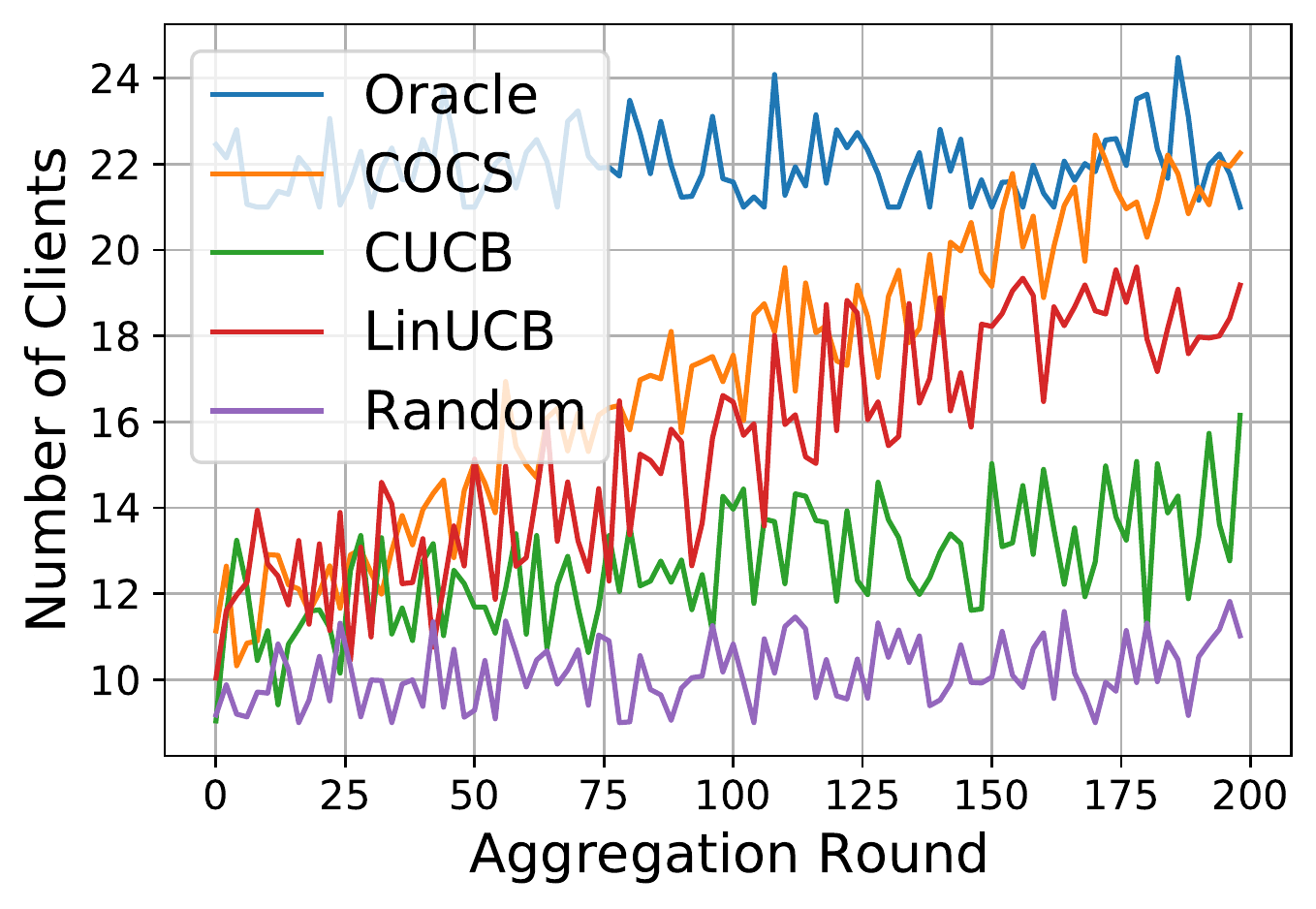}
  \subcaption{}
  \label{Fig:clients}
\end{minipage}
\hfill
\begin{minipage}{0.32\columnwidth}
  \vspace*{\fill}
  \centering
  \includegraphics[draft=false,width=\linewidth, height=0.72\linewidth]{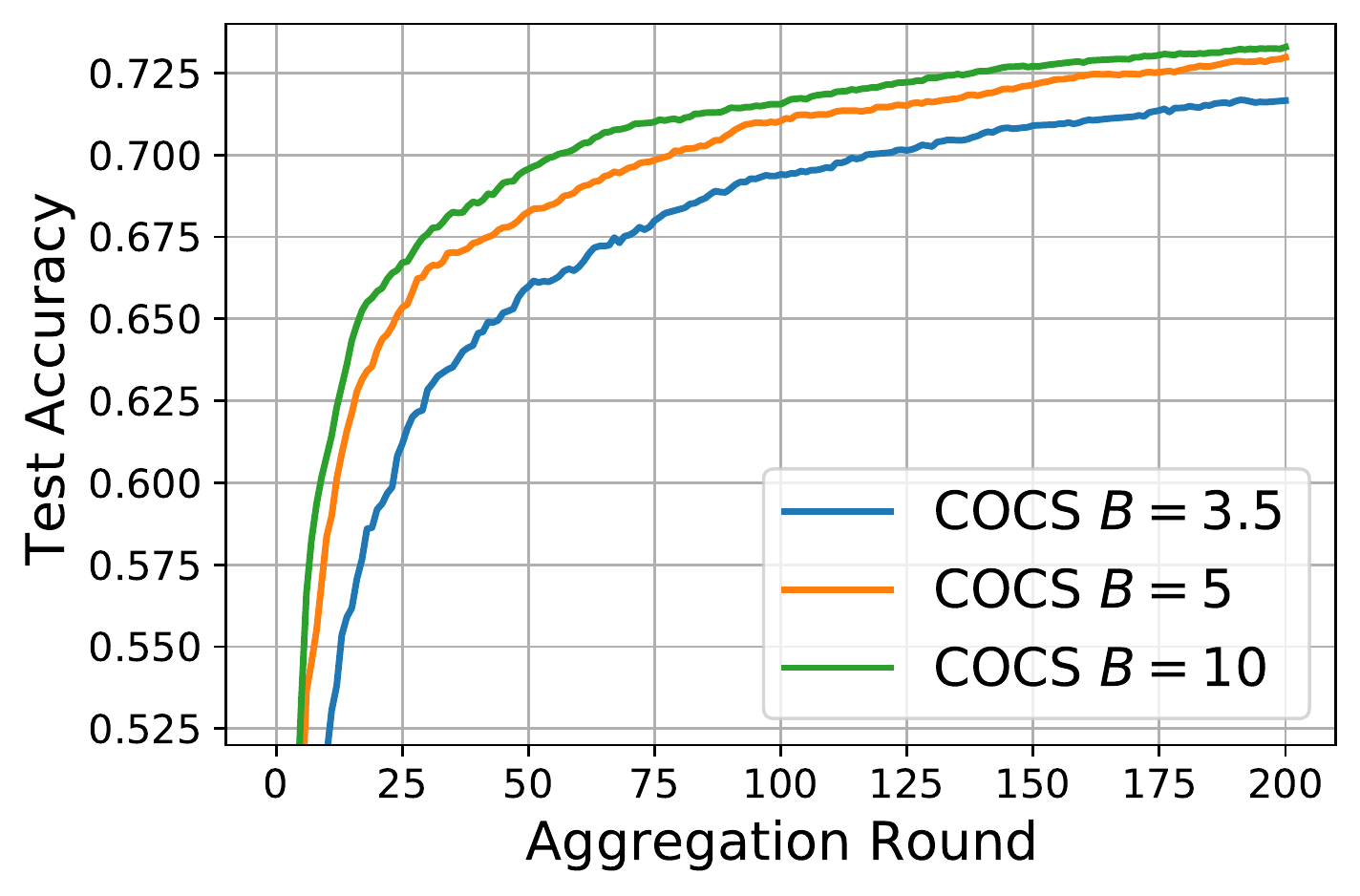}
  \subcaption{}
  \label{Fig:mnist_performanceB}\par\vfill
  \includegraphics[draft=false,width=\linewidth, height=0.72\linewidth]{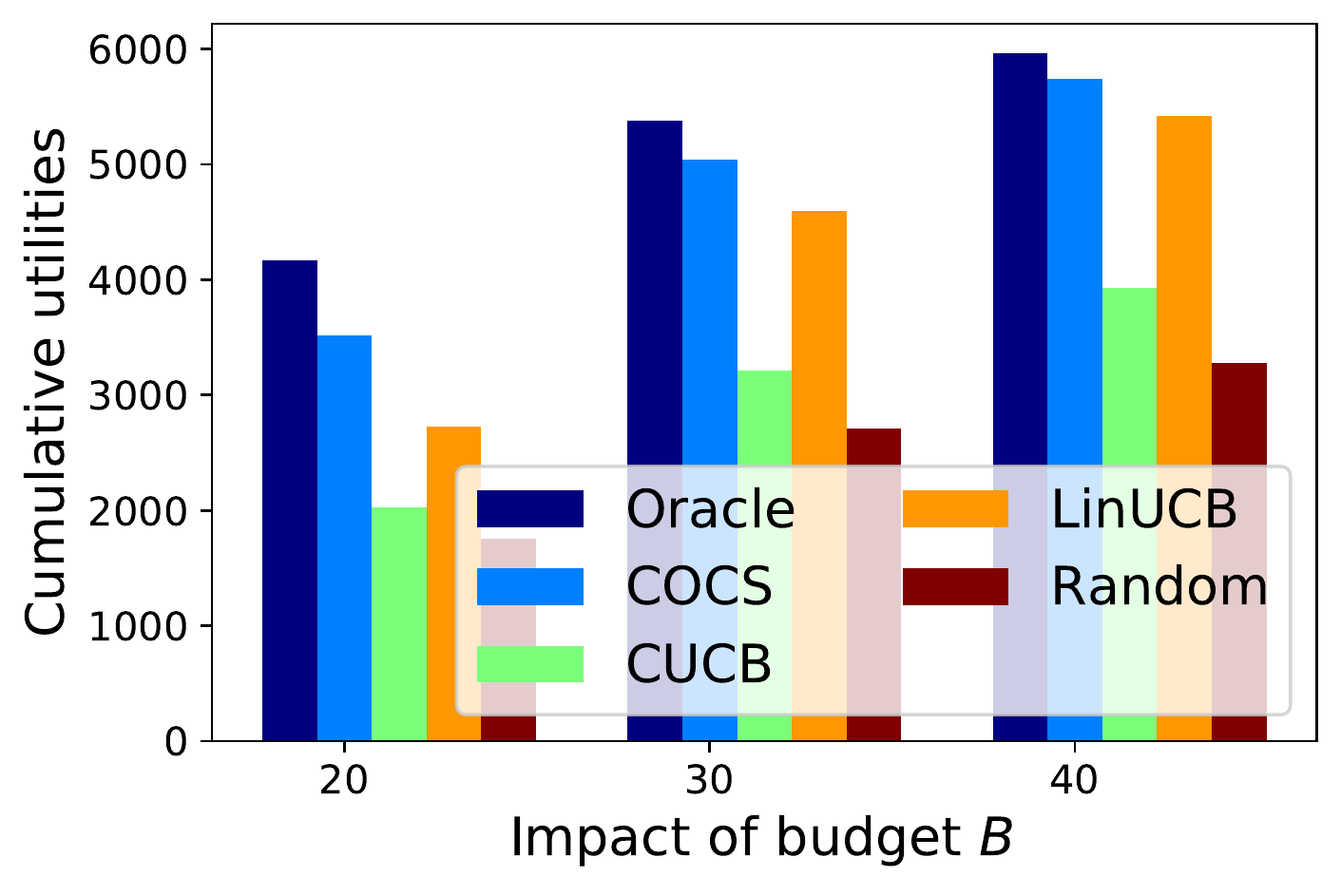}
  \subcaption{}
  \label{Fig:clientsB}
\end{minipage}
\hfill
\begin{minipage}{0.32\columnwidth}
  \vspace*{\fill}
  \centering
  \includegraphics[draft=false,width=\linewidth, height=0.72\linewidth]{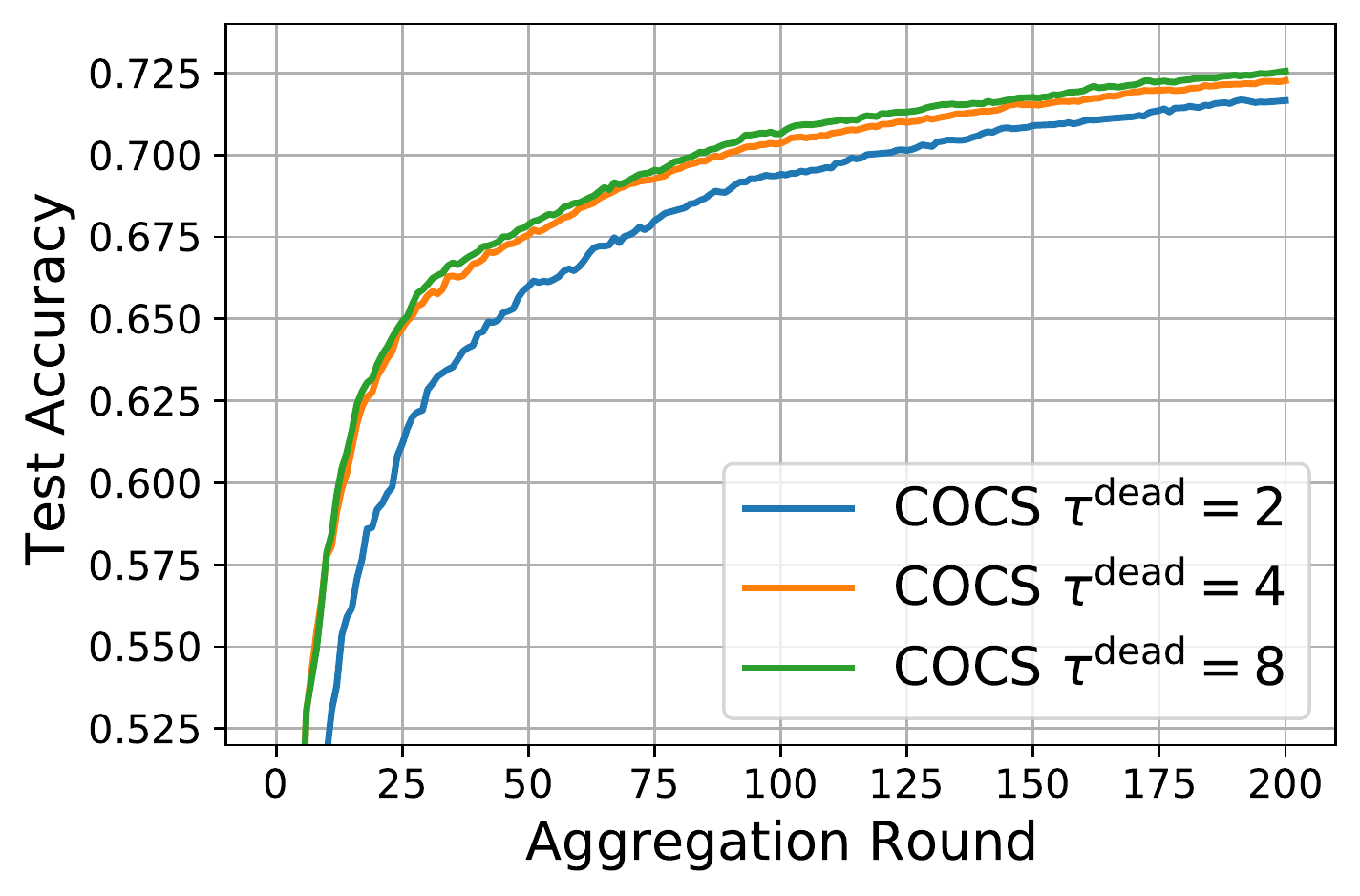}
  \subcaption{}
  \label{Fig:mnist_performancedead}\par\vfill
  \includegraphics[draft=false,width=\linewidth, height=0.72\linewidth]{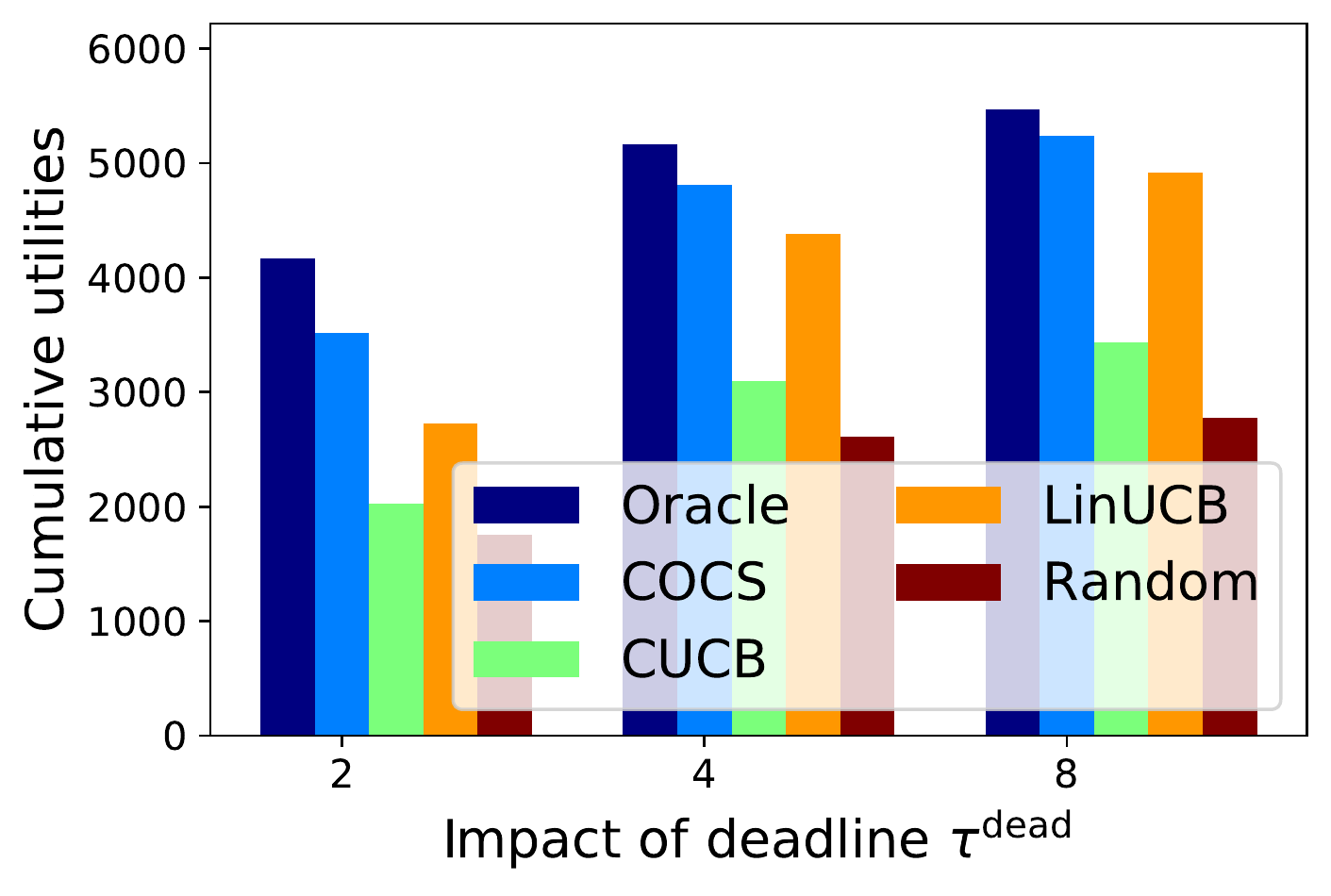}
  \subcaption{}
  \label{Fig:clientsdead}
\end{minipage}
\caption{(a) Training performance of 5 client selection policies based on logistic regression; (b) Temporal number of successful participated clients in each edge aggregation round (i.e., $\sum_{m\in\mathcal{M}}\sum_{n\in\bm{s}_m^t}X_{n,m}^t =1$); (c) Training performance of different budget $B$; (d) Cumulative utilities of different budget $B$; (e) Training performance of different deadline $\tau^{\text{dead}}$ and (f) Cumulative utilities of different deadline $\tau^{\text{dead}}$.}
\end{figure*}

\textbf{2) Training performance and client selection results:} We use two metrics to evaluate the training performance based on client selection policies: edge aggregation rounds to achieve targeted testing accuracy and final testing accuracy. In Fig.~\ref{Fig:mnist_performance}, we present the training performance under logistic regression on the MNIST dataset from different client selection benchmarks. Oracle policy, as expected, results in the fastest convergence speed and highest accuracy among all benchmarks. Although COCS performs slower than Oracle policy during the first several rounds due to exploration, it achieves similar testing accuracy to Oracle in 200th round. In particular, it easy to observe that COCS outperforms others. Due to the insufficient selection of clients all the rounds, Random selecting clients is considerably inferior to all other benchmarks. For clarifying the training performance, we present an auxiliary Table~\ref{Tab:performance} to emphasize the results, in which the targeted accuracy on MNIST dataset is set 70\%. As is shown, our proposed COCS policy only uses 121 rounds to achieve 70\% test accuracy, which is 36, 13 and 40 rounds faster than CUCB, LinUCB and Random.

In Fig.~\ref{Fig:clients}, we show the temporal number of clients are selected in each edge aggregation round. The upper bound and lower bound of clients are Oracle and Random policies. Although COCS, LinUCB, and CUCB all have increasing levels due to obtaining historical experiences, it is easy to observe that the increase of CUCB is very slow, and COCS outperforms the other two benchmarks. The reason why the number of successful participated clients are few in the first several rounds via COCS is that most of selected clients cannot be received by ESs before deadline $\tau^{\text{dead}}$. The results match the accumulative utilities and regret in Fig.~2.

\textbf{3) Impact of budget $B$:} Fig.~\ref{Fig:mnist_performanceB} shows that the training performance of COCS under different budgets $B$ ($B= 3.5, 5$ and $10$). It is easy to observe that COCS has better performance when NO increases the budget $B$. This is simply to explain because increasing the budget can select more clients in each round in order to increase the utility of HFL. In particular, when $B=5$ NO only uses 77 rounds to achieve 70\% test accuracy, which is much faster than $B=3.5$. However, the training performance does not have significant improvement from $B=5$ to $10$. Fig.~\ref{Fig:mnist_performanceB} presents the cumulative utilities of these five benchmarks after 200 edge aggregation rounds. Clearly, for all the benchmarks, the NO can achieve higher cumulative utilities with increasing budget. As can be observed, when the budget increases from $3.5$ to $5$, the benefit of client selection gradually increases. However, if the budget is set large, the redundant selection does not have significant improvement, which is similar to the results in Fig.~\ref{Fig:mnist_performanceB}. The reason may be because some client-ES pairs performing poor transmission rate cannot be successfully received by ESs before the deadline $\tau^{\text{dead}}$, even if NO increases the rental computation resources.

\textbf{4) Impact of deadline $\tau^{\text{dead}}$:} Fig.~\ref{Fig:mnist_performancedead} depicts the training performance and \ref{Fig:clientsdead} depicts the cumulative utilities under different deadlines $\tau^{\text{dead}} = 2$sec, $4$sec and $8$sec. Clearly, we can see that when NO increases the value of deadline, the number of clients increases gradually, which performs similarly to increasing budget $B$. However, if NO sets the deadline too large (i.e., $\tau^{\text{dead}}=8$sec), training performance and cumulative utilities perform similarly to $\tau^{\text{dead}}=4$sec. More specifically, cumulative utilities of COCS only increase from 5,125 to 5,378. Moreover, the increasing level of different $\tau^{\text{dead}}$ is less than increasing budget $B$, since the less budget can control the number of selected clients, which is more dominant than deadline for the training performance.

\begin{figure*}[t!]
  \centering
  \begin{minipage}{0.32\columnwidth}
      \includegraphics[draft=false,width=\linewidth, height=0.78\linewidth]{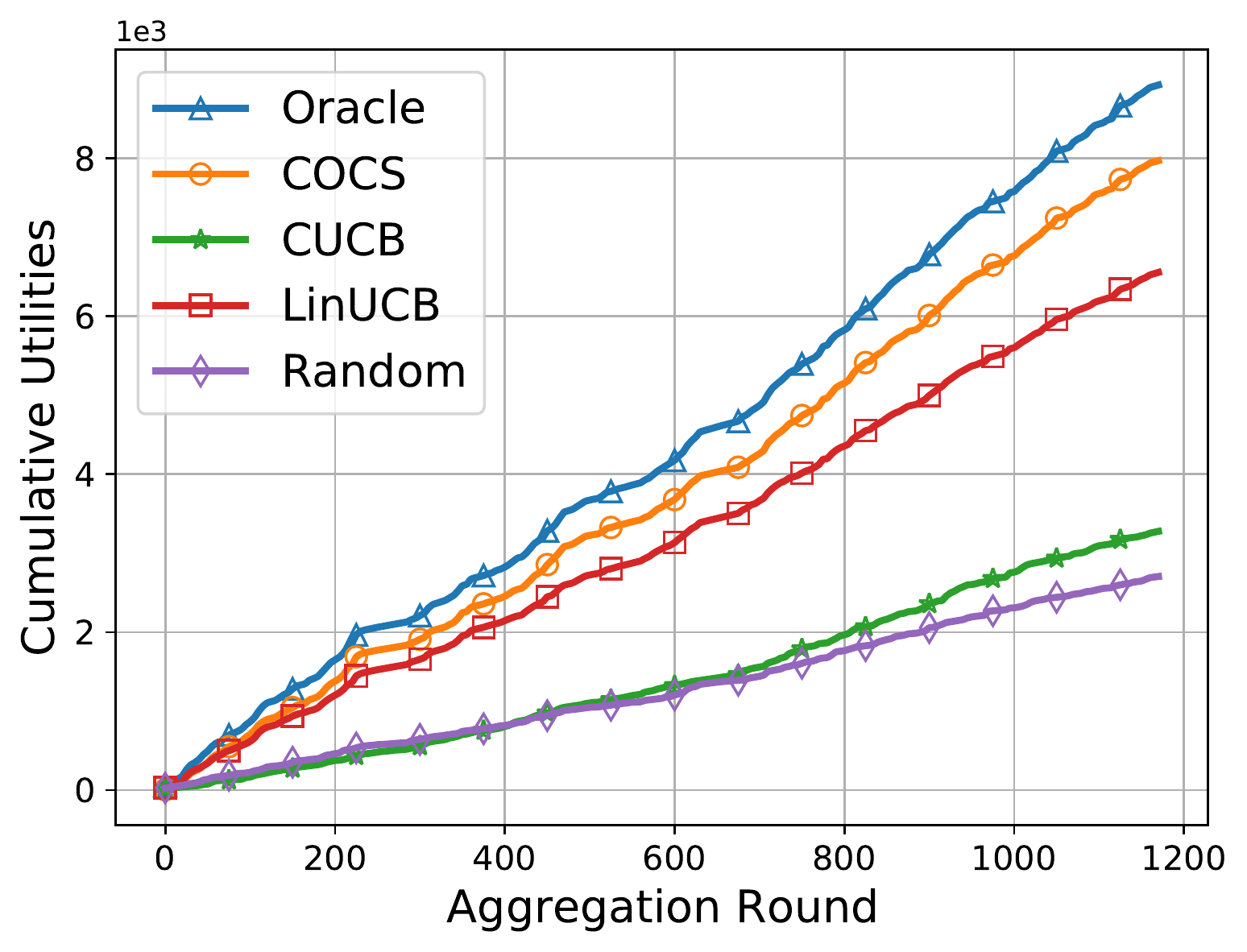}
      \caption{Cumulative Utilities under CNN on CIFAR-10 dataset.}
      \label{Fig:CIFArutility}
    \end{minipage}
    \hfill
    \begin{minipage}{0.32\columnwidth}
      \includegraphics[draft=false,width=\linewidth, height=0.78\linewidth]{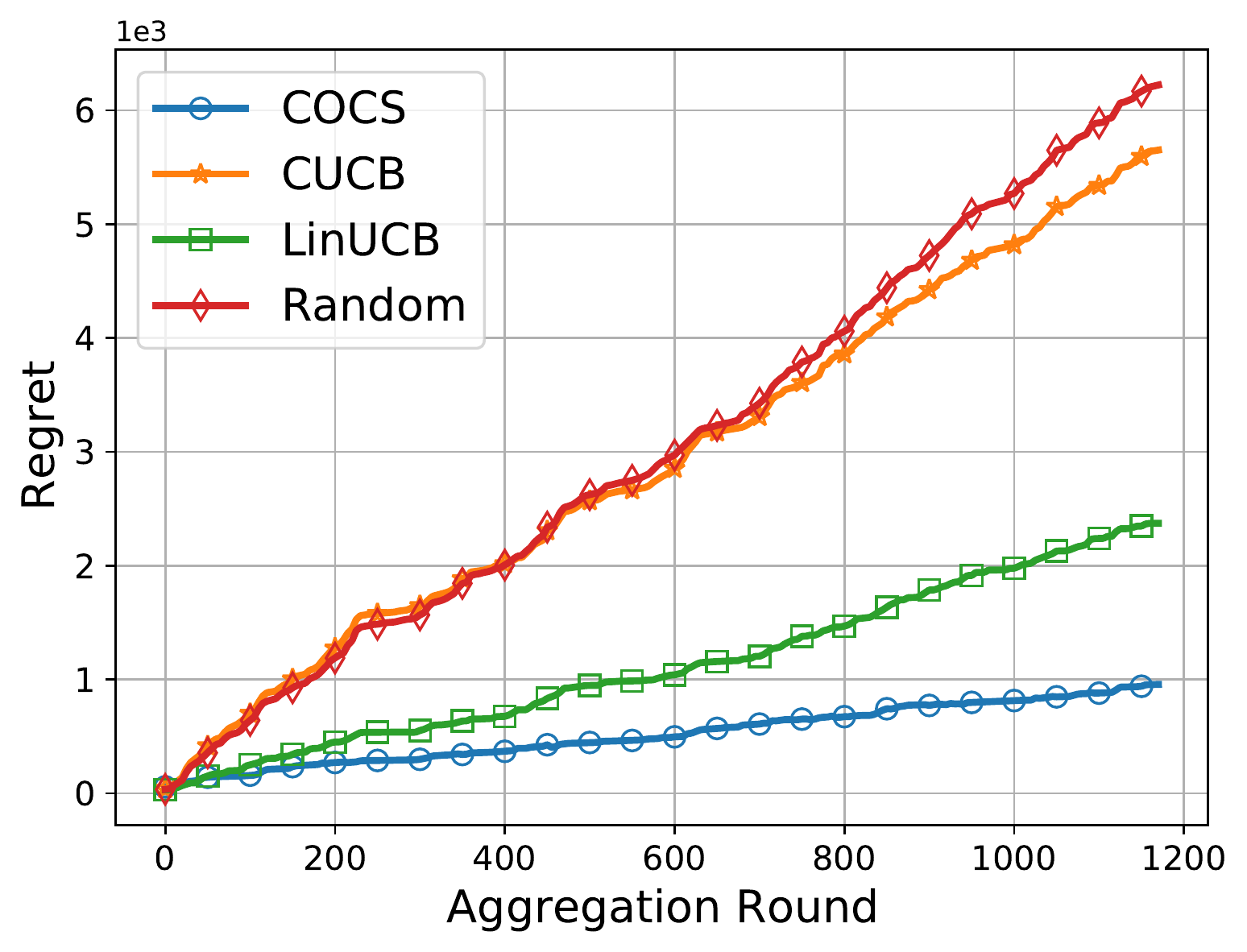}
      \caption{Regret under CNN on CIFAR-10 dataset.}
      \label{Fig:CIFARregret}
      \end{minipage}
      \hfill
    \begin{minipage}{0.32\columnwidth}
      \includegraphics[draft=false,width=\linewidth, height=0.78\linewidth]{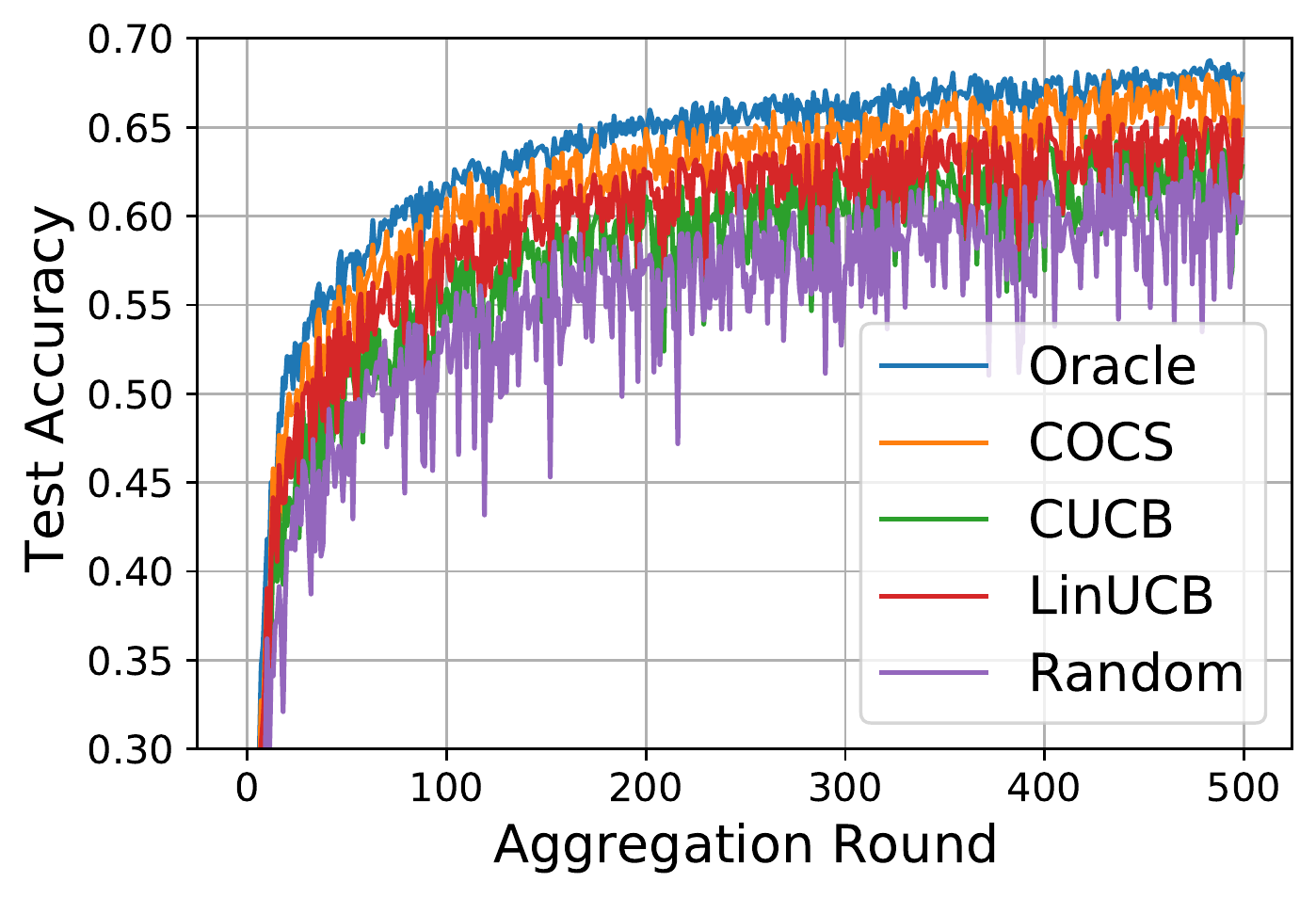}
      \caption{Training performance under CNN on CIFAR-10 dataset.}
      \label{Fig:CIFAR10}
    \end{minipage}
\end{figure*}

\begin{table}[t!]
\centering
\caption{Training performance of different benchmarks.}
\begin{tabular}{ccccc}
\hline
\multirow{2}{*}{Policy} & \multicolumn{2}{c}{MNIST (70\%)} & \multicolumn{2}{c}{CIFAR-10 (60\%)} \\ \cline{2-5} 
                        & Final Acc.         & Round        & Final Acc.          & Round          \\ \hline
Oracle                  & 71.90              & 111          & 68.41               & 92             \\ \hline
COCS                    & 71.84              & 121          & 67.93               & 101            \\ \hline
CUCB                     & 70.94              & 134          & 63.76               & 175            \\ \hline
LinUCB                  & 71.66              & 156          & 65.19               & 133            \\ \hline
Random                  & 70.81              & 161          & 62.25               & 207            \\ \hline
\end{tabular}
\label{Tab:performance}
\end{table}

\subsection{Performance Evaluation of Non-convex HFL}
In this subsection, we aim to show the performance of non-convex HFL under the CNN model on the CIFAR-10 dataset, where the utility is quadratically related to the number of participated clients. We set the error parameter of FLGreedy algorithm $\epsilon = 0.3$. The HFL network also includes 50 clients and 3 ESs, and the other parameters are introduced in Table~\ref{Tab:parameters}. Fig.~\ref{Fig:CIFArutility} depicts the cumulative utilities during 1,880 edge aggregation rounds and Fig.~\ref{Fig:CIFARregret} depicts regret. Similar to the performance of strongly convex HFL in Fig.~2, Oracle policy performs the best cumulative utilities as expected, and COCS outperforms the other 3 benchmarks, e.g., 1.7$\times$ higher than CUCB policy. In particular, the difference of cumulative utilities between Oracle and COCS is smaller than the result in Fig.~2a, since this is an approximated Oracle solution in this case. It is also observed that COCS achieves a sublinear regret in Fig.~\ref{Fig:CIFARregret}.

Fig.~\ref{Fig:CIFAR10} shows the test accuracy of different client selection benchmarks on the CIFAR-10 dataset. Since the training model and data size of CIFAR-10 is complicated enough, training performance of different benchmarks are clear to see. Oracle policy has the best performance among all benchmarks, which achieves 68.41\% test accuracy. In Table~\ref{Tab:performance}, COCS policy can achieve 67.93\% test accuracy, which is 4.17\%, 1.74\%, and 5.68\% higher than LinUCB, CUCB, and Random policies. From the convergence performance perspective, COCS also outperforms the other three benchmarks. We can conclude that COCS policy can improve the training performance for HFL significantly both on strongly convex and non-convex settings.

\section{Conclusion}\label{Sec:Conclusion}
In this paper, we investigated the client selection problem to improve the training performance for HFL. An online decision-making policy, called Context-aware Online Client Selection (COCS), was designed for NO to make proper client selection decisions for each client-ES pair. COCS was developed based on the CC-MAB framework, where NO observes the context of client-ES pairs and learns the participated probabilities to select clients and guide rental computation resources. Our proposed COCS policy departs from conventional optimization-based algorithms, which is able to work in HFL networks with uncertain information. More specifically, COCS addresses many practical challenges for HFL networks, and it is easy to implement and achieves a provably asymptotically optimal performance guarantee. Although our COCS policy has presented superior performance in extensive HFL experiments, there are still several future research questions. For example, considering dynamic partition of context space may improve the selection results, since it can generate more appropriate hypercubes. A theoretical improvement of convergence speed from COCS should also be considered.

\bibliographystyle{IEEEtran}

\bibliography{references}  

\clearpage

\appendices
\section{Proof of Lemma~\ref{lemma:boundexplore}}
\textit{Proof.} Firstly, we assume that the round $t$ is an exploration phase, based on the proposed COCS policy, at least one client $n$ has under-explore ESs, i.e., $\mathcal{L}_n^{\text{ue},t} \neq \emptyset$. As such, a hypercube $l_{n,m}^t$ of client $n$ satisfies $C_{n,m}^t (l)< K(t) = t^z \log(t)$. It is easy to see that there are at most $\lceil T^z \log(T) \rceil$ exploration phases, where computation resources of client $n$ are rented by the NO if its context satisfies $\phi_{n,m}^t \in l$. In each of these exploration phase, Let $\Psi^{\text{max},t}\triangleq \max_{\bm{s}, \bm{s}' \in \mathcal{S}^t} |\mu^t (\bm{s}; \bm{X}^t )-\mu^t (\bm{s}' ;\bm{X}^t )|$ denote the maximum utility loss for making a wrong client selection decision. Since the ideal decision is that all the clients participate in each aggregation round, and thus $\Psi^{\text{max},t}$ can be bounded by $N$. Let $c^{\text{min}}=\min_{n \in \mathcal{N}}c_n (y_n )$, the maximum number of clients can be rented with $f_n^t >0$ can be bounded by $BN/c^{\text{min}}$. Therefore, the regret incurred in one aggregation round is bounded by $BN/c^{\text{min}}$. Due to the fact that there are at most $\lceil K(T) \rceil$ exploration phases for a hypercube $l$ of a client-ES pair $(n,m)$ until to $T$ rounds, $h_T$ hypercubes in the partition $\mathcal{L}_T$, and $NM$ client-ES pairs, then $\mathbb{E}[R_{\text{explore}}(T)]$ can be bounded by:
\begin{equation}
    \mathbb{E}[R_{\text{explore}}(T)] \leq \frac{N^2 MB}{c^{\text{min}}}(h_T)^2 \lceil T^z \log(T) \rceil = \frac{N^2 MB}{c^{\text{min}}}\lceil T^\gamma \rceil^2 \lceil T^z \log(T) \rceil.\nonumber
\end{equation}
Using $\lceil T^\gamma \rceil^2 \leq 4T^{2\gamma} = 4 T^{2\gamma}$, we have:
\begin{equation}\label{Eq:boundRexplore}
    \mathbb{E}[R_{\text{explore}}(T)] \leq \frac{N^2 MB}{c^{\text{min}}}4 T^{2\gamma} ( T^z \log(T) + 1) = \frac{N^2 MB}{c^{\text{min}}}4 ( T^{z+2\gamma} \log(T) + T^{2\gamma}).
\end{equation}
The proof is completed. \hfill~$\Box$

\section{Proof of Lemma~\ref{lemma:boundexploit}}
\textit{Proof.} Before proving the Lemma~\ref{lemma:boundexploit}, we first define some auxiliary variables. For a hypercube $l \in \mathcal{L}_T$, we define $\bar{p} (l) = \sup_{\phi\in l}p (\phi)$ and $\underline{p} (l) = \inf_{\phi\in l}p (\phi)$ denote the best and worst expected quality of all contexts $\phi \in l$. Therefore, given the PC set $\mathcal{S}^t = \{1,2,\dots, S^t \}$, context set $\bm{\phi}^t = \{\phi_1^t ,\phi_2^t ,\dots, \phi_{S^t}^t\}$, and the hypercube set $\bm{l}^t = \{l_1^t , l_2^t ,\dots, l_{S^t}^t\}$ for each client-ES pair in aggregation round $t$. Let
\begin{equation}
\begin{split}
    \bar{\bm{p}}(\bm{l}^t ) & = \{\bar{p}(l^t_1 ) ,\bar{p} (l^t_2 ),\dots, \bar{p} (l_{S^t}^t) )\}\\
    \underline{\bm{p}}(\bm{l}^t ) & = \{\underline{p} (l^t_1 ),\underline{p} (l^t_2 ),\dots, \underline{p} (l_{S^t}^t) )\}\nonumber
\end{split}
\end{equation}
Note that we need to compare the context qualities at different positions in a hypercube $l$ in some proof steps. Therefore, we define the context at the center of a hypercube $l$ as $\dot{\phi}_l$ and its expected quality $\dot{p}(l) = p(\dot{\phi}_l )$. Let $\dot{\bm{p}}^t = \{\dot{p} (l_1^t), \dot{p} (l_2^t ), \dots, \dot{p}(l_{S^t}^t)\}$. We also define the client selection decision $\dot{\bm{\mu}}(\bm{s}^t ; \dot{\bm{p}}(\bm{l}^t ))$, which is derived from the expected PPs $\dot{\bm{p}}(\bm{l}^t)$ by solving the following optimization in edge aggregation round $t$:
\begin{equation}
    \dot{\bm{s}}^t = \argmax_{\bm{s}^t \in\mathcal{S}^t } \mu^t (\bm{s}^t ;\dot{\bm{p}}(\bm{l}^t ))~~~~\text{s.t.~~\eqref{Eq:singlebudget}, \eqref{Eq:singlebudget}}
\end{equation}
The client selection decision $\dot{\bm{\mu}}(\bm{l}^t )$ identifies the bad selection decisions when the hypercubes of contexts $\bm{\phi}^t$ is $\bm{l}^t$. Let
\begin{equation}
    \mathcal{P}^t = \{G \in \mathcal{S}^t | \mu(\dot{\bm{s}}^t; \underline{p} (\bm{l}^t )) - \mu(G; \bar{\bm{p}}(\bm{l}^t )) \geq At^\theta\}
\end{equation}
denote the set of suboptimal client selection decision when the contexts of client-ES pair are included into $\bm{l}^t$. The parameter $A >0$ and $\theta <0$ are for the regret analysis. Recall that a client selection decision $G \in \mathcal{P}(\bm{l}^t)$ is suboptimal for $\bm{l}^t$, since the utility of NO achieved by the client selection decision $\dot{\bm{s}}^t$ is at least $At^\theta$, and it is higher than the sum of the best number of PCs selection for subset $G$. The client selection decision in $\mathcal{S}^t \setminus \mathcal{P}(\bm{l}^t )$ is called near-optimal for $\bm{l}^t$. Therefore, the regret of $R_{\text{exploit}}(T)$ can be divided into the following two summands:
\begin{equation}
    \mathbb{E}[R_{\text{exploit}}(T)] = \mathbb{E}[R_s (T)] + \mathbb{E}[R_n (T)]
\end{equation}
where the term $\mathbb{E}[R_s (T)]$ is the regret due to the suboptimal client selection decision in exploitation phase and $\mathbb{E}[R_n (T)]$ is the regret due to the near-optimal client selection decision in exploitation phase. Next, we will demonstrate that each of the two summands can be bounded. Firstly, we present the bound of $\mathbb{E}[R_s (T)]$ in Lemma~\ref{lemma:Rs}.

\begin{lemma}\label{lemma:Rs}
(Bound of $\mathbb{E}[R_s (T)]$.) Given the input parameters $K(t)=t^z \log(t)$ and $h_T = \lceil T^\gamma \rceil$, where $0 < z <1$ and $0 < \gamma < \frac{1}{2}$, if the H\"{o}lder condition holds and the additional condition $2H(t)+\frac{2NMB}{c^{\text{min}}}L2^{\frac{\alpha}{2}}h_T^\alpha \leq At^\theta$ is satisfied with $H(t) = \frac{NMB}{c^{\text{min}}}t^{-\frac{z}{2}}$ for all $t$, then $\mathbb{E}[R_s (T)]$ is bounded by:
\begin{equation}
    \mathbb{E}[R_s (T)] \leq\frac{BNM}{c^{\text{min}}}\bigg(\sum_{k=1}^{B/c^{\text{min}}}\begin{pmatrix}
N\\
k
\end{pmatrix}\bigg)\frac{\pi^2}{3} , \nonumber
\end{equation}
where $\sum_{k=1}^{B/c^{\text{min}}}\begin{pmatrix}
N\\
k
\end{pmatrix}$ is the maximum possible number of client-ES pairs with the size less than or equal the budget constraint $B/c^{\text{min}}$.
\end{lemma}

\textit{Proof.} For the aggregation round $1\leq t \leq T$, let $W^{t}(t) = \mathcal{C}^{\text{ue},t}_n = \emptyset$ denote the event that COCS policy enters the exploitation phase. According to the definition of $\mathcal{C}^{\text{ue},t}_n$, we have $C_{n,m}^t (l^t )> K(t), \forall n , \forall l^t \in \bm{l}^t$. Let $S_G^t$ be the event that the subset of clients $G \in \mathcal{P}^t$ is selected in edge aggregation round $t$. Then, we have
\begin{equation}
    R_s (T) = \sum_{t=1}^{T}\sum_{G \in\mathcal{P}^t } I_{\{S_G^t , W^t \}}\times (\mu (\bm{s}^{\text{opt},t}; \bm{X}^{t})- \mu (G^t;\bm{X}^t )).
\end{equation}
In each summand, the utility loss comes from a suboptimal decision $G \in\mathcal{P}^t$ instead of the optimal Oracle decision $\bm{s}^{\text{opt},t}$. Due to the fact that the maximum utility loss of NO can be bounded by $NM$, and the maximum number of clients can be selected due to the budget constraint is $BN/c^{\text{min}}$, we have:
\begin{equation}
    R_s (T) \leq \frac{N^2MB}{c^{\text{min}}}\sum_{t=1}^{T}\sum_{G\in\mathcal{P}^t }I_{\{S_G^t, W(t) \}}.
\end{equation}
Calculating the expectation, the regret from suboptimal decision can be bounded by:
\begin{equation}
    \mathbb{E}[R_s (T)] \leq \frac{N^2 MB}{c^{\text{min}}} \sum_{t=1}^{T}\sum_{G \in\mathcal{P}^t }\mathbb{E}\bigg[I_{\{S_G^t ,W(t) \}}\bigg] = \frac{N^2 MB}{c^{\text{min}}} \sum_{t=1}^{T}\sum_{G\in\mathcal{P}^t}\mathbb{P}\{S_G^t , W(t) \}.
\end{equation}
Based on the COCS policy, we know that if a client selection decision $\bm{s}^t$ enters in exploration phase (i.e., the event $V_S^t$), the rewards of NO is at least as high as the reward of clients which are selected in $\dot{\bm{s}}^t$, i.e., $\mu(G ; \hat{\bm{X}}^t ) \geq \mu(\dot{\bm{s}}^t ; \hat{\bm{X}}^t )$. Therefore, we have:
\begin{equation}\label{Eq:PVW}
    \mathbb{P}\{V_S^t , W^t \} \leq \mathbb{P}\{\mu (G ;\hat{\bm{X}}^t)) \geq \mu(\dot{\bm{s}}^t ; \hat{\bm{X}}^t ),W(t)\}.
\end{equation}
The right-hand side of Eq.~\eqref{Eq:PVW} implies at least one of the three following events with any $H(t)>0$:
\begin{equation}
\begin{split}
    E_1 &= \{\mu(G ;\hat{\bm{X}}^t ) \geq \mu (\bar{\bm{p}}^t ; G)+ H(t), W(t)\},\\
    E_2 &= \{\mu (\dot{\bm{s}}^t ; \hat{\bm{X}}^t ) \leq \mu (\dot{\bm{s}}^t ;\underline{\bm{p}}^t) - H(t),W(t)\},\\
    E_3 &= \{G ;\hat{\bm{X}}^t ) \geq \mu (\dot{\bm{s}}^t; \hat{\bm{X}}^t); \mu(G ;\hat{\bm{X}}^t ) < \mu(G; \bar{\bm{p}}^t ) + H(t), \mu(\dot{\bm{s}}^t ;\hat{\bm{X}}^t ) > \mu(\dot{\bm{s}}^t ;\underline{\bm{p}}^t )-H(t),W(t)\}.
\end{split}
\end{equation}
Hence, the original event is Eq.~\eqref{Eq:PVW} can be shown as follows:
\begin{equation}
    \{\mu (G; \hat{\bm{X}}^t ) \geq \mu(\dot{\bm{s}}^t ;\hat{\bm{X}}^t ), W(t)\} \subseteq \{E_1 \cup E_2 \cup E_3 \}.
\end{equation}
We will separately bound the probability of these three events $E_1$, $E_2$ and $E_3$. Firstly, we aim to bound the event $E_1$. Recall that the best expected quality of client selections for the hypercube $l$ is $\bar{p}(l) = \sup_{\phi\in l}p (\phi)$. Thus, the expected PP of client $n$ in $G$ is bounded as follows:
\begin{equation}
    \mathbb{E}[\hat{p}_n^t (l)] = \mathbb{E}\bigg[\frac{1}{C^t_n (l)}\sum_{l \in \mathcal{E}_n^t (l)}X(\phi_n^t )\bigg] \leq \bar{p}_n^t (l).
\end{equation}
This yields
\begin{equation}
\begin{split}
    \mathbb{P}\{E_1 \} & = \mathbb{P}\{\mu (G; \hat{\bm{X}}^t )\geq \mu (G; \bar{\bm{p}}^t +H(t) ),W(t)\}\\
    & = \mathbb{P}\bigg\{\hat{X} (l_{n,m}^t ) \geq \bar{p}_n (l_{n,m}^t )+\frac{H(t) c^{\text{min}}}{BN} , \exists (n,m) \in G , W(t)\bigg\}\\
    & = \mathbb{P}\bigg\{\hat{X} (l_{n,m}^t ) \geq \mathbb{E}[\hat{X}_n (l_{n,m}^t )]+\frac{H(t) c^{\text{min}}}{BN} , \exists (n,m) \in G , W(t)\bigg\}\\
   & \leq \sum_{(n,m) \in G}\mathbb{P}\bigg\{\hat{X}_n (l_{n,m}^t ) \geq \mathbb{E}[\hat{X}_n (l_{n,m}^t )] +\frac{H(t)c^{\text{min}}}{BN}, W(t)\bigg\}.
\end{split}
\end{equation}

Based on Chernoff-Hoeffding bound \cite{hoeffding1994probability}, we claim that for each client, the upper-bound of the PC estimation is $1$ and exploiting that event $W^t$ implies that at least $t^z \log(t)$ samples were drawn, then we have:
\begin{equation}
\begin{split}
    \mathbb{P}\{E_1 \} & \leq \sum_{(n,m)\in G}\mathbb{P} \bigg\{\hat{X}_n (l_{n,m}^t ) - \mathbb{E}[\hat{X}_n (l_{n,m}^t )] \geq \frac{H(t) c^{\text{min}}}{BN}, W(t) \bigg\}\\
    & \leq \sum_{(n,m) \in G}\exp \bigg(-2C_{n,m}^t (l) H(t)^2 \bigg(\frac{c^{\text{min}}}{BN}\bigg)^2 \bigg) \\
    & \leq \sum_{(n,m) \in G}\exp \bigg(-2 H(t)^2 t^z \log(t) \bigg(\frac{c^{\text{min}}}{BN}\bigg)^2\bigg).
\end{split}
\end{equation}
Similarly, event $E_2$ can be proven as follows:
\begin{equation}
    \mathbb{P}\{E_2 \} \leq \mathbb{P} \bigg\{\mu(\dot{\bm{s}}^t ; \hat{\bm{X}}^t ) \geq \mu(\dot{\bm{s}}^t ; \underline{\bm{p}}^t ) - H(t), W(t) \bigg\} \leq \sum_{(n,m)\in\dot{\bm{s}}^t}\exp\bigg(-2 H(t)^2 t^z \log(t) \bigg(\frac{c^{\text{min}}}{BN}\bigg)^2\bigg).
\end{equation}
In order to bound the event $E_3$, we first define some auxiliary parameters. First, we rewrite the estimated PP $\hat{p}_n (l), l \in \mathcal{L}_T$ as follows:
\begin{equation}
    \hat{p}_n = \frac{1}{C_n^t (l)} \sum_{(t' , n' ,m'): \phi_{n,m}^{t'}\in l , (n,m)\in \bm{s}^{t'}}p(\phi_{n,m}^{t'}) + \xi_n^{t'} ,\nonumber
\end{equation}
where $\xi_n^{t'}$ is the deviation from estimation $\hat{p}_n (\phi_{n,m}^{t'} )$ from the expected PC $n$ with context $\phi_{n,m}^{t'}$. In additional, the best and worst context in the hypercube $l \in \mathcal{L}^T$ are defined as $\phi^{\text{best}}(l)=\argmax_{\phi\in l} p(\phi)$ and $\phi^{\text{worst}}(l)=\argmin_{\phi \in l} p(\phi)$. Let
\begin{equation}
\begin{split}
    p^{\text{best}}(l) & = \frac{1}{C_n^t (l)} \sum_{(t' , n' ,m'): \phi_{n,m}^{t'}\in l , (n,m)\in \bm{s}^{t'}} p(\phi_{n,m}^{\text{best}}(l)) + \xi_n^{t'} \\
    p^{\text{worst}}(l) & = \frac{1}{C_n^t (l)} \sum_{(t' , n' ,m'): \phi_{n,m}^{t'}\in l , (n,m)\in \bm{s}^{t'}} p(\phi_{n,m}^{\text{worst}}(l)) + \xi_n^{t'} .\nonumber
\end{split}
\end{equation}
Let $\bm{p}^{\text{best},t} = [\beta^{\text{best}}(l_{1,1}^t ), \dots, p^{\text{best}}(l_{S^t}^t )]$ and $\bm{p}^{\text{worst},t} = [p^{\text{worst}}(l_{1,1}^t ), \dots, p^{\text{worst}}(l_{S^t}^t )]$. Based on the H\'{o}lder condition, due to the fact that $\phi^{\text{best}}(l) \in l$ and the contexts from hypercube $l$ are used for calculating the estimated PCs $\hat{X}(l)$, it can be shown that
\begin{align}
    p^{\text{best}}(l_{n,m}^t ) - \hat{p}(l^t_{n,m} ) & \leq L2^{\frac{\alpha}{2}}h_T^{-\alpha} \label{Eq:best}\\
    \hat{p}(l_{n,m}^t ) - p^{\text{worst}}(l_{n,m}^t ) & \leq L2^{\frac{\alpha}{2}}h_T^{-\alpha} .\label{Eq:worst}
\end{align}
Applying the above results, we will have
\begin{align}
    & \mu(G; \bm{p}^{\text{best},t} ) - \mu (G; \hat{\bm{X}}^t ) \leq \sum_{(n,m) \in G}(p^{\text{best}}(l_{n,m}^t) - \hat{p}(l_{n,m}^t )) \leq \frac{BN}{c^{\text{min}}}L2^{\frac{\alpha}{2}}h_T^{-\alpha}\\
    & \mu(\dot{\bm{a}}^t; \hat{\bm{X}}^t ) - \mu (\dot{\bm{a}}^t; \bm{p}^{\text{worst},t}) \leq \sum_{(n,m) \in \dot{\bm{p}}^t}(\hat{p}(l_{n,m}^t )- p^{\text{worst},t} (l_{n,m}^t)) \leq \frac{BN}{c^{\text{min}}}L2^{\frac{\alpha}{2}}h_T^{-\alpha},
\end{align}
Based on the definition of $p^{\text{best}}(l)$ and $p^{\text{worst}}(l)$, it holds for the first part of $E_3$ that
\begin{equation}
    E_{\text{3.1}} = \{\mu (G; \hat{\bm{X}}^t ) \geq \mu (\dot{\bm{s}}^t ; \hat{\bm{X}}^t )\} \subseteq \{\mu(G; \bm{p}^{\text{best},t}) \geq \mu (\dot{\bm{s}}^t ; \bm{p}^{\text{worst},t} \}
\end{equation}
For the second part, using Eq.~\eqref{Eq:best}, we have
\begin{equation}
\begin{split}
    E_{\text{3.2}} & = \{\mu(G; \hat{\bm{X}}^t ) < \mu(G; \bar{\bm{p}}^t + H(t)\}\\
    & \subseteq \bigg\{\mu(G; \bm{p}^{\text{best},t}) + \frac{BN}{c^{\text{min}}}L2^{\frac{\alpha}{2}}h_T^{-\alpha} < \mu(G; \bar{\bm{p}}^t ) + H(t)\bigg\}\\
    & = \bigg\{\mu(G; \bm{p}^{\text{best},t}) < \mu(G; \bar{\bm{p}}^t ) + \frac{BN}{c^{\text{min}}}L2^{\frac{\alpha}{2}}h_T^{-\alpha} + H(t)\bigg\}
\end{split}
\end{equation}
For the third part, using Eq.~\eqref{Eq:worst}, we have
\begin{equation}
\begin{split}
    E_{\text{3.3}} & = \{\mu (\dot{\bm{s}}^t ; \hat{\bm{X}}^t ) > \mu(\dot{\bm{s}}^t ; \underline{\bm{p}}^t ) - H(t)\}\\
    & \subseteq \bigg\{\mu (\dot{\bm{s}}^t ; \bm{p}^{\text{worst},t}) + \frac{BN}{c^{\text{min}}}L2^{\frac{\alpha}{2}}h_T^{-\alpha} > \mu (\dot{\bm{s}}^t ;\underline{\bm{p}}^t - H(t)\bigg\}\\
    & = \bigg\{\mu (\dot{\bm{s}}^t ; \bm{p}^{\text{worst},t}) > \mu(\dot{\bm{s}}^t ; \underline{\bm{p}}^t) - \frac{BN}{c^{\text{min}}} L2^{\frac{\alpha}{2}}h_T^{-\alpha} - H(t)\bigg\}
\end{split}
\end{equation}

We aim to find a condition under which the probability of $E_3$ is zero. To this end, we design the parameter $H(t)$, such that
\begin{equation}\label{Eq:Ht}
    2H(t) + \frac{2BN}{c^{\text{min}}}L2^{\frac{\alpha}{2}}h_T^{-\alpha} \leq At^\theta .
\end{equation}
Since $G \in \mathcal{P}^t $, we have $\mu(\dot{\bm{s}}^t ; \underline{\bm{p}}^t ) - \mu(G ;\bar{\bm{p}}^t ) \geq At^\theta$. Combining with the results in Eq.~\eqref{Eq:Ht} we have:
\begin{equation}
    \mu (\dot{\bm{s}}^t ; \underline{\bm{p}}^t ) - \mu(G ;\bar{\bm{p}}^t ) - 2H(t) - \frac{2BN}{c^{\text{min}}}L2^{\frac{\alpha}{2}}h_T^{-\alpha} \geq 0 .
\end{equation}
Rewriting yields that:
\begin{equation}\label{Eq:rewrite}
    \mu (\dot{\bm{s}}^t ; \underline{\bm{p}}^t ) -H(t) - \frac{BN}{c^{\text{min}}} L2^{\frac{\alpha}{2}}h_T^{-\alpha} \geq \mu(G ;\bar{\bm{p}}^t + H(t) + \frac{BN}{c^{\text{min}}} L2^{\frac{\alpha}{2}}h_T^{-\alpha} .
\end{equation}
If Eq.~\eqref{Eq:rewrite} is true, the three parts of $E_3$ cannot be satisfied at the same time: combining $E_{\text{3.2}}$ and $E_{\text{3.3}}$ with Eq.~\eqref{Eq:rewrite} yields $\mu(G; \bm{p}^{\text{best},t}) < \mu(\dot{\bm{s}}^t ; \bm{p}^{\text{worst},t})$, which contradicts the $E_{\text{3.1}}$. Therefore, under the condition in Eq.~\eqref{Eq:Ht}, we have the results that $\mathbb{P}\{E_3 \}=0$. Until now, the analysis is based on an arbitrary $H(t)>0$. In the remainder of the proof, we choose $H(t) = \frac{BN}{c^{\text{min}}}t^{-z/2}$. Then, we have
\begin{equation}
    \mathbb{P}\{E_1\} \leq \frac{BN}{c^{\text{min}}} \exp\bigg(-2H(t)^2 t^z \log(t)\bigg(\frac{c^{\text{min}}}{BN}\bigg)^2 \bigg) \leq \frac{BN}{c^{\text{min}}} \exp(-2\log(t)) \leq \frac{BN}{c^{\text{min}}} t^{-2}.
\end{equation}
Similarly, 
\begin{equation}
    \mathbb{P}\{E_2 \} \leq \frac{BN}{c^{\text{min}}}t^{-2}.
\end{equation}

To sum up, under condition Eq.~\eqref{Eq:Ht}, the probability $\mathbb{P}\{S_G^t, W(t)\}$ can be bounded by:
\begin{equation}
    \mathbb{P}\{S_G^t , W(t)\} \leq \mathbb{P}\{E_1 \cup E_2 \cup E_3 \} \leq \mathbb{P}\{E_1\} +\mathbb{P}\{E_2 \} + \mathbb{P} \{E_3 \} \leq \frac{2BN}{c^{\text{min}}}t^{-2}.
\end{equation}
Based on the above analysis, the regret for $\mathbb{E}[R_s (T)]$ can be bounded as follows:
\begin{equation}
\begin{split}
    \mathbb{E}[R_s (T)] & \leq \frac{BN}{c^{\text{min}}} \sum_{t=1}^{T} \sum_{G \in \mathcal{L}^t} \mathbb{P}\{S_G^t , W(t)\} \leq \frac{BN}{c^{\text{min}}} \sum_{t=1}^{T}|\mathcal{L}^t | \frac{2BN}{c^{\text{min}}}t^{-2} \leq \frac{BN}{c^{\text{min}}}|\mathcal{L}^t |2 \sum_{t=1}^{T}t^{-2} \\
    & \leq \frac{BN}{c^{\text{min}}}|\mathcal{L}^t |\frac{\pi^2}{3} \leq \frac{BNM}{c^{\text{min}}}\bigg(\sum_{k=1}^{B/c^{\text{min}}}\begin{pmatrix}
N\\
k
\end{pmatrix}\bigg)\frac{\pi^2}{3} .
\end{split}
\end{equation}
where $\sum_{k=1}^{B/c^{\text{min}}}\begin{pmatrix}
N\\
k
\end{pmatrix}$ is the maximum possible number of clients on each ES with the size less than or equal the budget constraint $B/c^{\text{min}}$. \hfill~$\Box$

Next, we bound the regret due to choosing near-optimal client selection decisions, which is given in the following Lemma.
\begin{lemma}\label{lemma:Rn}
(Bound for $\mathbb{E}[R_n (T)]$). Given the input parameters $h_T= \lceil T^\gamma \rceil$ and $K(t) = t^z \log(t)$ , where $0 <z<1$ and $0 < \gamma < \frac{1}{2}$, if the H\"{o}lder condition holds true, the regret $\mathbb{E}[R_n (T)]$ is bounded by:
\begin{equation}
    \mathbb{E}[R_n (T)] \leq \frac{3NMB}{c^{\text{min}}} L2^{\frac{\alpha}{2}} T^{1-\gamma \alpha} + \frac{A}{1+\theta}T^{1+\theta}.
\end{equation}
\end{lemma}

\textit{Proof.} For an arbitrary edge aggregation round $t$, let the event $W(t)$ denotes as the same in Lemma~\ref{lemma:Rs}, the regret is due to the near optimal client selection decision is:
\begin{equation}
    R_n (T) = \sum_{t=1}^T  I_{\{\bm{s}^t \in \{\mathcal{S}^t \setminus \mathcal{P}(\bm{l}^t ) \}, W(t) \}} \times (\mu(\bm{s}^{\text{opt},t} ;\bm{p}^t ) - \mu(\bm{s}^t ; \bm{p}^t )),
\end{equation}
where the selected clients subset $\bm{s}^t$ is near-optimal in each time slot, i.e., $\bm{s}^t \in \mathcal{S}^t \setminus \mathcal{P}^t$, the regret is considered for selecting $\bm{s}^t$ not the $\bm{s}^{\text{opt},t}$. Let $Q (t) = W (t) \cap \{\bm{s}^t \in \mathcal{S}^t \setminus \mathcal{P}(\bm{l}^t )\}$ denote the event of selecting a near-optimal arm set. Taking the expectation of $R_n (T)$, by the definition of conditional expectation, we have:
\begin{equation}
\begin{split}
    \mathbb{E}[R_n (T)] & = \sum_{t=1}^T \mathbb{P}\{Q(t) \} \cdot \mathbb{E}[\mu(\bm{a}^{\text{opt},t} ;\bm{p}^t ) - \mu(\bm{s}^t ; \bm{p}^t ) | Q(t) ]\\
    & \leq \sum_{t=1}^T \mathbb{E}[\mu(\bm{s}^{\text{opt},t} ;\bm{p}^t ) - \mu(\bm{s}^t ; \bm{p}^t ) |Q(t) ].
\end{split}
\end{equation}
If the COCS policy enter an exploitation phase and $\mathcal{K} \in \mathcal{S}^t \setminus \mathcal{P}(\bm{l}^t )$, based on the definition of $\mathcal{L}^{\text{ue},t}_n$, it holds that $C^t_{n,m} > K(t) = t^z \log(t), \forall l_{n,m}^t \in \bm{l}^t$. Since $\mathcal{K} \in \mathcal{S}^t \setminus \mathcal{P}(\bm{l}^t )$, we have the following results:
\begin{equation}
    \mu(\dot{\bm{s}}^t ; \underline{\bm{p}}^t ) - \mu(\mathcal{K} ; \bar{\bm{p}}^t ) < At^\theta.
\end{equation}
To bound the regret, we have to give an upper bound on:
\begin{equation}
    \sum_{t=1}^T \mathbb{E}[\mu(\bm{s}^{\text{opt},t} ;\bm{p}^t ) -\mu(\bm{s}^t ;\bm{p}^t )|Q(t)] = \sum_{t=1}^T \mu(\bm{a}^{\text{opt},t} ;\bm{p}^t )-\mu(\mathcal{K};\bm{p}^t ) .
\end{equation}
After applying H\"{o}lder condition several times, we have
\begin{equation}
\begin{split}
    \mu(\bm{s}^{\text{opt},t} ;\bm{p}^t ) - \mu(\bm{s}^t ; \bm{p}^t ) & \leq \mu(\bm{s}^{\text{opt},t} ; \dot{\bm{s}}^t ) + \frac{BN}{c^{\text{min}}}L2^{\frac{\alpha}{2}}h_T^{-\alpha} - \mu (\mathcal{K} ; \bm{p}^t )\\
    & \leq \mu(\dot{\bm{s}}^t ; \underline{\bm{p}}^t ) + \frac{BN}{c^{\text{min}}}L2^{\frac{\alpha}{2}}h_T^{-\alpha} - \mu (\mathcal{K} ; \bm{p}^t )\\
    & \leq \mu(\dot{\bm{s}}^t ; \underline{\bm{p}}^t ) + \frac{3BN}{c^{\text{min}}}L2^{\frac{\alpha}{2}}h_T^{-\alpha} - \mu (\mathcal{K} ; \bar{\bm{p}}^t )\\
    & \leq \frac{3BN}{c^{\text{min}}} L2^{\frac{\alpha}{2}}h_T^{-\alpha} + At^\theta \leq \frac{3BN}{c^{\text{min}}} L2^{\frac{\alpha}{2}} T^{-\gamma \alpha} + At^\theta .
\end{split}
\end{equation}
Since there are $M$ ESs, $\mathbb{E}[R_n (T)]$ can be bounded by:
\begin{equation}
    \mathbb{E}[R_n (T)] \leq \sum_{t=1}^{T} \bigg(\frac{3NMB}{c^{\text{min}}} L2^{\frac{\alpha}{2}} T^{-\gamma \alpha} + At^\theta \bigg) \leq \frac{3NMB}{c^{\text{min}}} L2^{\frac{\alpha}{2}} T^{1-\gamma \alpha} + \frac{A}{1+\theta}T^{1+\theta} .\nonumber
\end{equation}
Proof is completed. \hfill~$\Box$

Therefore, the regret $\mathbb{E}[R_{\text{exploit}}(T)]$ is bounded by:
\begin{equation}
    \mathbb{E}[R_{\text{exploit}}(T)] \leq \frac{BNM}{c^{\text{min}}}\bigg(\sum_{k=1}^{B/c^{\text{min}}}\begin{pmatrix}
N\\
k
\end{pmatrix}\bigg)\frac{\pi^2}{3}+ \frac{3NMB}{c^{\text{min}}} L2^{\frac{\alpha}{2}} T^{1-\gamma \alpha} + \frac{A}{1+\theta}T^{1+\theta} .\nonumber
\end{equation}
The proof is completed. \hfill~$\Box$

\section{Proof of Theorem~\ref{theorem:RT}}
\textit{Proof.} Let $K(t) = t^z \log(t)$, $0 < z<1$ and $h_T = \lceil T^\gamma \rceil$, $0<\gamma < \frac{1}{2}$, $0 < \gamma < \frac{1}{D}$, $H(t) = \frac{NMB}{c^{\text{min}}}t^{-\frac{z}{2}}$, and the condition $2H(t) + \frac{2NMB}{c^{\text{min}}}L2^{\frac{\alpha}{2}}T^{-\gamma \alpha} \leq At^{\theta}$ is satisfied for all $t$. Combining Lemmas~\ref{lemma:boundexplore} and \ref{lemma:boundexploit}, the regret $R(T)$ can be bounded by:
\begin{equation}
\begin{split}
     \mathbb{E}[R(T)] & = \mathbb{E}[R_{\text{explore}}(T)] + \mathbb{E}[R_{\text{exploit}}(T)]\\ 
   & \leq \frac{N^2 MB}{c^{\text{min}}}4 (T^{z+2\gamma}\log(T) + T^{2\gamma})\\
   & + \frac{NMB}{c^{\text{min}}}\bigg(\sum_{k=1}^{B/c^{\text{min}}}\begin{pmatrix}
N\\
k
\end{pmatrix}\bigg)\frac{\pi^2}{3}\\
& + \frac{3NMB}{c^{\text{min}}} L2^{\frac{\alpha}{2}} T^{1-\gamma \alpha} + \frac{A}{1+\theta}T^{1+\theta}.
\end{split}
\end{equation}
The summands contribute to the regret $R(T)$ due to the leading orders $O(\log (T)T^{z+2\gamma})$, $O(T^{1-\gamma \alpha})$ and $O(T^{1+\theta})$. By selecting the parameters $z, \gamma , A$ and $\theta$ as $z = \frac{2\alpha}{3\alpha +2} \in (0,1)$, $\gamma = \frac{z}{2\alpha}$, $\theta = -\frac{z}{2}$, and $A = \frac{2NMB}{c^{\text{min}}} + \frac{2NMB}{c^{\text{min}}}L2^{\frac{\alpha}{2}}$, we can balance the leading orders. Therefore, the regret $R(T)$ reduces to
\begin{equation}\label{Eq:leadingbound}
\begin{split}
    \mathbb{E}[R(T)] & \leq \frac{4N^2 MB}{c^{\text{min}}}(\log(T)T^{\frac{2\alpha + 2}{3\alpha +2}} + T^{\frac{2}{3\alpha + 2}})\\
    & + \frac{NMB}{c^{\text{min}}}\bigg(\sum_{k=1}^{B/c^{\text{min}}}\begin{pmatrix}
N\\
k
\end{pmatrix}\bigg)\frac{\pi^2}{3}\\
& +\bigg(\frac{3NMB}{c^{\text{min}}} L2^{\frac{\alpha}{2}} + \frac{2NMB}{c^{\text{min}}}\frac{1+L2^{\frac{\alpha}{2}}}{(2\alpha + 2)/(3\alpha + 2)}\bigg)T^{\frac{2\alpha + 2}{3\alpha +2}}.
\end{split}
\end{equation}
From the bound in Eq.~\eqref{Eq:leadingbound}, the dominant order is 
\begin{equation}
O\bigg(\frac{4N^2 MB}{c^{\text{min}}}T^{\frac{2\alpha +D}{3\alpha +D}}\log (T)\bigg).
\end{equation}
The proof is completed. \hfill~$\Box$

\section{Proof of Theorem~\ref{The:submodular}}\label{SubSec:submodular}
\noindent\textit{Proof.} Since we have presented knapsack and matroid constraints from \eqref{Eq:singlebudget} and \eqref{Eq:singlematroid} in Section~\ref{SubSec:oracle}, it only remains to show that \textbf{P3} is monotone submodular maximization. It is easy to verify the monotone property, as square function is monotone increasing. Consider any sets $A \subseteq B \subseteq \mathcal{N} \times \mathcal{M}$, and the client-ES pair $(n' ,m') \notin B$, then we have:
\begin{equation}
\begin{split}
    &\bigg(\frac{1}{M}\sum_{(n,m)\in A \cup (n',m')}p_{n,m}^t \bigg)^{\frac{1}{2}} - \bigg(\frac{1}{M}\sum_{(n,m)\in A}p_{n,m}^t \bigg)^{\frac{1}{2}}\\
    & \approx \frac{p_{n',m'}^t}{2(M\sum_{(n,m) \in A}p_{n,m}^t )^\frac{1}{2}} \geq \frac{p_{n',m'}^t}{2(M\sum_{(n,m) \in B}p_{n,m}^t )^\frac{1}{2}} \\
    &\approx \bigg(\frac{1}{M}\sum_{(n,m)\in B \cup (n',m')}p_{n,m}^t \bigg)^{\frac{1}{2}} - \bigg(\frac{1}{M}\sum_{(n,m)\in B}p_{n,m}^t \bigg)^{\frac{1}{2}}, \nonumber
\end{split}
\end{equation}
where the above proof is based on the first order Taylor expansion \cite{roberts2014introductory}. Therefore, the client selection decision of non-convex HFL in \textbf{P3} is a submodular maximization. This concludes the proof. ~\hfill$\Box$

\section{Proof of Theorem~\ref{The:appbound}}\label{SubSec:appbound}
\noindent\textit{Proof.} The proof of Theorem~\ref{The:appbound} is similar to the proof of Theorem~\ref{theorem:upperbound}, and thus we only present a proof sketch of Theorem~\ref{The:appbound}. The expected $\delta$-regret is also divided into two parts:
\begin{equation}
    \mathbb{E}[R^{\delta}(T)] = \mathbb{E}[R^{\delta}_{\text{explore}}(T)] + \mathbb{E}[R^{\delta}_{\text{exploit}}(T)].
\end{equation}
Since the worst-case utility loss $\frac{MB}{c^{\text{min}}}$ is used to present a upper bound of the regret $\mathbb{E}[R^{\delta}_{\text{explore}}(T)]$ due to a wrong client selection decision, the approximation algorithm FLGreedy does not have much influences for bounding $\mathbb{E}[R^{\delta}_{\text{explore}}(T)]$. Based on the definition of $\delta$-regret, the worst-case utility loss is $\frac{1}{\delta}\frac{MB}{c^{\text{min}}}$. Based on the proof in Lemma~\ref{lemma:boundexplore}, $\mathbb{E}[R^{\delta}_{\text{explore}}(T)]$ can be bounded by:
\begin{equation}
    \mathbb{E}[R^{\delta}_{\text{explore}}(T)] \leq \frac{4N^2 MB}{\delta c^{\text{min}}}(T^{\frac{2\alpha +2}{3\alpha +2}}\log(T) + T^{\frac{2}{3\alpha+2}}).
\end{equation}
Now, we will provide a upper bound of $\mathbb{E}[R^{\delta}_{\text{exploit}}(T)]$ by identifying suboptimal and near-optimal set of arms:
\begin{equation}
    \mathcal{P}^\delta (\bm{l}^t ) = \{G \in \mathcal{S}^t | \mu(\dot{\bm{s}}^\delta (\bm{l}^t ); \underline{p} (\bm{l}^t )) - \mu(G; \bar{\bm{p}}(\bm{l}^t )) \geq At^\theta\},
\end{equation}
where $\mu(\dot{\bm{s}}^\delta (\bm{l}^t )$ is the $\delta$-approximation by solving the following optimization:
\begin{equation}
    \max_{\bm{s}^t} \mu (\bm{s}^t ;\dot{\bm{p}} (\bm{l}^t ))~~~\text{s.t.}~ \text{\eqref{Eq:singlebudget}, \eqref{Eq:singlematroid}}.
\end{equation}
We divide $\mathbb{E}[R^{\delta}_{\text{exploit}}(T)] = \mathbb{E}[R_s^\delta (T)] + \mathbb{E}[R_n^\delta (T)]$ into two parts, which can be bounded separately.

For the regret of suboptimal decisions in $\mathcal{P}^\delta (\bm{l}^t )$ in exploitation phase (i.e., $\mathbb{E}[R_s^\delta (T)]$), we have:
\begin{equation}
\begin{split}
    \mathbb{E}[R_s^\delta (T)] & = \sum_{t=1}^{T}\sum_{\bm{s}^t \in \mathcal{P}^\delta (\bm{l}^t )} \mathbb{E}\bigg[I_{\{S_G^t , W^t \}} \times (\frac{1}{\delta}\mu (\bm{s}^{\text{opt},t}; \bm{X}^{t})- \mu (G^t;\bm{X}^t ))\bigg]\\
    & \leq \frac{1}{\delta}\frac{N^2 MB}{c^{\text{min}}} \sum_{t=1}^{T}\sum_{\bm{s}^t \in \mathcal{P}^\delta (\bm{l}^t )}\mathbb{P}\{S_G^t , W(t)\}.
\end{split}
\end{equation}
Similar to the proof in Lemma~\ref{lemma:Rs}, we have:
\begin{equation}
    \mathbb{E}[R_s^\delta (T)] \leq \frac{NMB}{\delta c^{\text{min}}}\bigg(\sum_{k=1}^{B/c^{\text{min}}}\begin{pmatrix}
N\\
k
\end{pmatrix}\bigg)\frac{\pi^2}{3} 
\end{equation}
For the regret of choosing near-optimal decisions in $\S \setminus \mathcal{P}^\delta (\bm{l}^t )$ in exploitation phase (i.e., $\mathbb{E}[R_n^\delta (T)]$), we have:
\begin{equation}
\begin{split}
    \mathbb{E}[R_n^\delta (T)] & = \sum_{t=1}^{T}\mathbb{P}\{W(t), Q(t)\} \cdot \mathbb{E}\bigg[\frac{1}{\delta} \mu(\bm{s}^{\text{opt},t}; \bm{X}^t) - \mu(\bm{s}^t ; \bm{X}^t )| W(t), Q(t) \bigg]\\
    & \leq \sum_{t=1}^{T} \mathbb{E}\bigg[\frac{1}{\delta} \mu(\bm{s}^{\text{opt},t}; \bm{X}^t) - \mu(\bm{s}^t ; \bm{X}^t )| W(t), Q(t) \bigg].
\end{split}
\end{equation}
Since $\bm{s}^t \in \S \setminus \mathcal{P}^\delta (\bm{l}^t )$, it holds that:
\begin{equation}
    \mu (\dot{\bm{s}}^{\delta} (\bm{l}^t ) ; \underline{\bm{p}}(\bm{l}^t )) - \mu (\bm{s}^t ; \bar{\bm{p}}(\bm{l}^t )) < \frac{A}{1+\theta}T^{1+\theta} .
\end{equation}
To bound the regret, we present an upper bound as
\begin{equation}
    \sum_{t=1}^{T}\mathbb{E}\bigg[\frac{1}{\delta} \mu(\bm{s}^{\text{opt},t}; \bm{X}^t) - \mu(\bm{s}^t ; \bm{X}^t )| W(t), Q(t) \bigg] = \sum_{t=1}^{T}\frac{1}{\delta} \mu(\bm{s}^{\text{opt},t}; \bm{p}^t) - \mu(\bm{s}^t ; \bm{p}^t ) .
\end{equation}
Applying H\"{o}lder condition server times, we have:
\begin{equation}
\begin{split}
    \frac{1}{\delta} \mu(\bm{s}^{\text{opt},t}; \bm{p}^t) - \mu(\bm{s}^t ; \bm{p}^t ) & \leq \frac{1}{\delta} \mu(\bm{s}^{\text{opt},t}; \dot{\bm{p}}^t) + NMLD^{\frac{\alpha}{2}}h_T^{-\alpha} - \mu(\bm{s}^t ; \bm{p}^t )\\
    & \leq \mu(\dot{\bm{s}}^{\delta}(\bm{l}^t ); \dot{\bm{p}}^t) + \frac{NMB}{c^{\text{min}}}L2^{\frac{\alpha}{2}}h_T^{-\alpha} - \mu(\bm{s}^t ; \bm{p}^t )\\
    & \leq \mu(\dot{\bm{s}}^{\delta}(\bm{l}^t ); \underline{\bm{p}}^t) + \frac{2NMB}{c^{\text{min}}}L2^{\frac{\alpha}{2}}h_T^{-\alpha} - \mu(\bm{s}^t ; \bm{p}^t )\\
    & \leq \mu(\dot{\bm{s}}^{\delta}(\bm{l}^t ); \underline{\bm{p}}^t) + \frac{3NMB}{c^{\text{min}}}L2^{\frac{\alpha}{2}}h_T^{-\alpha} - \mu(\bm{s}^t ; \bm{p}^t )\\
    & \leq \frac{3NMB}{c^{\text{min}}}L2^{\frac{\alpha}{2}} T^{1-\gamma \alpha} + \frac{A}{1+\theta}T^{1+\theta} .
\end{split}
\end{equation}
Therefore, the regret $\mathbb{E}[R^{\delta}_{\text{exploit}}(T)]$ is bounded by:
\begin{equation}
    \mathbb{E}[R^{\delta}_{\text{exploit}}(T)] \leq \frac{NMB}{\delta c^{\text{min}}}\bigg(\sum_{k=1}^{B/c^{\text{min}}}\begin{pmatrix}
N\\
k
\end{pmatrix}\bigg)\frac{\pi^2}{3} + \frac{3NMB}{c^{\text{min}}}L2^{\frac{\alpha}{2}} T^{1-\gamma \alpha} + \frac{A}{1+\theta}T^{1+\theta}.
\end{equation}
Comparing $\mathbb{E}[R^{\delta}_{\text{explore}}(T)]$ and $\mathbb{E}[R^{\delta}_{\text{exploit}}(T)]$, we can see that the dominant orders of $\mathbb{E}[R^{\delta}(T)]$ is based on $\mathbb{E}[R^{\delta}_{\text{explore}}(T)]$ as follows
\begin{equation}
   O\bigg(\frac{4N^2 MB}{\delta c^{\text{min}}}T^{\frac{2\alpha +2}{3\alpha +2}}\log (T)\bigg).
\end{equation}
This concludes the proof. ~\hfill$\Box$

\end{document}